\documentclass{article}
\pdfoutput=1 

\PassOptionsToPackage{numbers,sort&compress}{natbib}

\usepackage[main,final]{neurips_2026}

\usepackage[utf8]{inputenc} 
\usepackage[T1]{fontenc}    
\usepackage{hyperref}       
\usepackage{url}            
\usepackage{booktabs}       
\usepackage{amsfonts}       
\usepackage{nicefrac}       
\usepackage{microtype}      
\usepackage{xcolor}         

\usepackage{amsmath,amssymb}
\usepackage{tabularx}
\usepackage{longtable}
\makeatletter
\let\inputrows\@@input
\makeatother
\usepackage{multirow}
\usepackage{graphicx}
\usepackage{subcaption}
\usepackage{float}
\usepackage{cleveref}
\usepackage{enumitem}
\usepackage{xspace}
\usepackage{listings}
\usepackage{pifont}
\newcommand{\cmark}{\textcolor{green!50!black}{\ding{51}}}
\newcommand{\xmark}{\textcolor{red!75!black}{\ding{55}}}
\newcommand{\spclass}{\textcolor{gray!70!black}{\ding{109}}}    
\newcommand{\itclass}{\textcolor{blue!55!black}{\ding{108}}}    

\lstdefinestyle{fragcal}{
  basicstyle=\ttfamily\scriptsize,
  keywordstyle=\bfseries\color{blue!60!black},
  stringstyle=\color{green!40!black},
  commentstyle=\itshape\color{gray},
  numberstyle=\ttfamily\tiny\color{gray},
  numbers=left,
  numbersep=6pt,
  tabsize=4,
  breaklines=true,
  breakatwhitespace=true,
  columns=fullflexible,
  keepspaces=true,
  showstringspaces=false,
  frame=single,
  rulecolor=\color{gray!40},
  framexleftmargin=4pt,
  xleftmargin=10pt,
  upquote=true,
  literate=
    {—}{{--}}1
    {–}{-}1
    {─}{-}1,
  aboveskip=6pt,
  belowskip=6pt,
}

\hypersetup{
  colorlinks=true,
  linkcolor=blue!70!black,
  citecolor=blue!70!black,
  urlcolor=blue!70!black,
  bookmarksnumbered=true,
}

\crefname{table}{Table}{Tables}
\crefname{figure}{Figure}{Figures}
\crefname{section}{Section}{Sections}
\crefname{equation}{Eq.}{Eqs.}
\crefname{appendix}{Appendix}{Appendices}

\title{When Calibration Data Matter in Sub-4-Bit Language Model Quantization}

\author{%
  Yehong Jiang\textsuperscript{1,2}\thanks{Corresponding author: Yehong Jiang. This work was conducted during his internship at Intel Corporation.},\quad
  Yen-Kuang Chen\textsuperscript{2},\quad Xinmin Tian\textsuperscript{2} \\
  \textsuperscript{1}Massachusetts Institute of Technology \quad
  \textsuperscript{2}Intel Corporation \\
  \texttt{\{\href{mailto:yehong.jiang@intel.com}{yehong.jiang},
  \href{mailto:y.k.chen@intel.com}{y.k.chen},
  \href{mailto:xinmin.tian@intel.com}{xinmin.tian}\}@intel.com};\quad
  \href{mailto:yehong28@mit.edu}{\texttt{yehong28@mit.edu}}
}

\begin{document}
\maketitle

\begin{abstract}
Low-bit post-training quantization reduces language-model memory but can sharply degrade downstream task accuracy.
Calibration usually relies on generic text, but when task-informed data recover accuracy and which quantizers benefit remain poorly understood.
We introduce FragCal-DC, which supplements generic text with training documents from task families with low retention of full-precision accuracy in a two-bit scan.
We use a two- to four-bit sweep to locate functional regimes and four-seed calibration comparisons to evaluate six data-dependent methods across six dense models (1.7B--8B) and ten tasks.
At their lowest tested functional bit widths, iterative SignRound~V2 and OmniQuant improve at two bits, whereas single-pass GPTQ and AWQ show small mean responses at three bits.
At two bits, SignRound~V2 gains $1.83$ percentage points in mean ten-task accuracy over generic-text calibration across six models.
At 70 billion parameters, GPTQ and SignRound~V2 both improve at two bits, with functional generic-calibration baselines.
For two-bit SignRound~V2, task-level controls show that recovery concentrates on families included in calibration, while incorrect answer designations retain about 90\% of the mean accuracy gain.
Changes in reconstruction error do not consistently track accuracy gains.
These findings support task-informed calibration as a design principle for preserving downstream accuracy in low-bit quantization.

\end{abstract}

\section{Introduction}
\label{sec:intro}

Post-training quantization (PTQ) reduces language-model memory by representing trained weights with fewer bits, but can degrade task accuracy. Leading data-dependent PTQ methods use calibration data, sample text that guides quantization, typically from general-purpose corpora~\citep{frantar2023gptq,lin2024awq,shao2024omniquant,cheng2025signroundv2,egiazarian2024aqlm}. Calibration composition is the choice and mixture of these samples. Activation-aware Weight Quantization (AWQ) and OmniQuant report limited changes in perplexity, a measure of next-token prediction quality, across calibration sources~\citep{lin2024awq,shao2024omniquant}. Broader quantization and pruning comparisons find task- and method-dependent accuracy variation~\citep{williams2024calibration}. We ask when data from downstream tasks recover accuracy below four bits per weight, which quantizers benefit, and where recovery occurs.

We introduce \emph{FragCal-DC}, a task-informed calibration mixture of benchmark training data and general-purpose text from the Pile corpus~\citep{gao2020pile}. An exploratory five-model two-bit scan identifies fragile task families by low ratios of quantized to unquantized (full-precision) accuracy. We vary calibration composition while keeping each quantizer's update rule fixed (\cref{sec:setup}).

\begin{figure}[t]
\centering
\includegraphics[width=\linewidth]{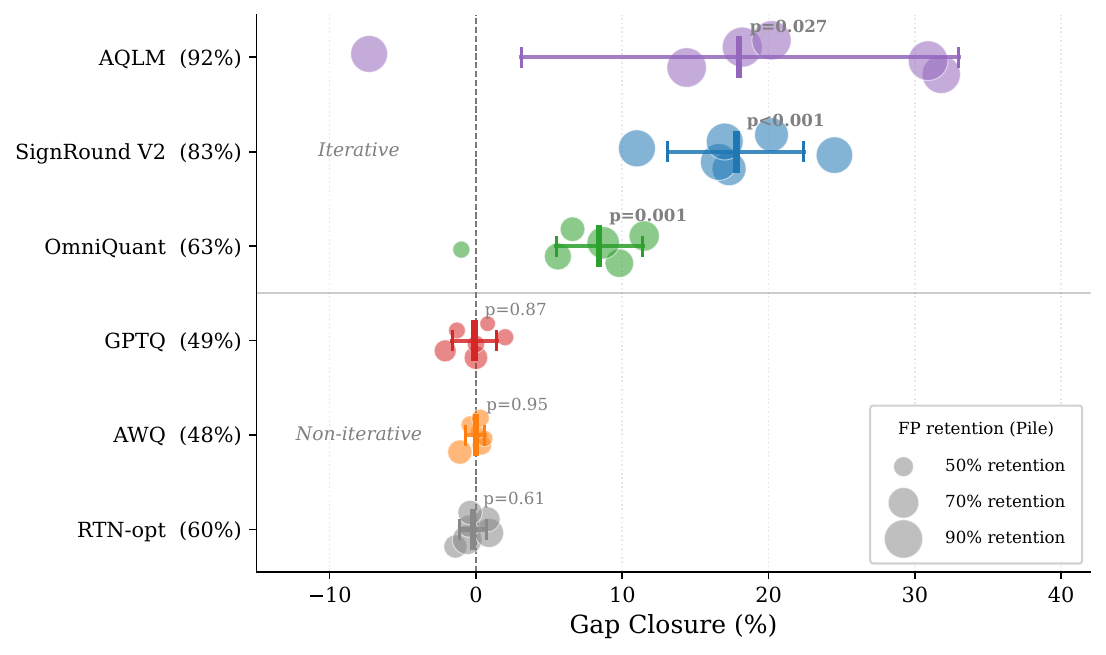}
\caption{Calibration responses at a common two-bit budget.
Each bubble shows one model's gap closure, the percentage of the gap between generic-calibration
and full-precision accuracy recovered by FragCal. Bubble area encodes full-precision retention
(generic/full-precision accuracy).
Ticks and whiskers show across-model means and 95\% confidence intervals
($n=6$ models, OmniQuant $n=5$).
OmniQuant's Mistral-7B bubble is excluded from summaries.
AQLM uses approximately two bits per weight before overhead.
See \cref{sec:setup} for test details.}
\label{fig:method_type}
\end{figure}

At two bits, changing calibration text recovers a larger mean percentage of the generic-calibration accuracy gap to full precision for methods with repeated gradient-based updates than for those without them (\cref{fig:method_type}). We call methods with these updates iterative and those without them single-pass. However, the single-pass GPTQ/AWQ baselines are largely collapsed, defined here as generic-calibration mean ten-task accuracy at or below $0.35$. Above this threshold, configurations are functional.

We therefore compare calibration responses at functional operating widths, at a common three-bit width, and on a larger model. A two- to four-bit sweep locates functional configurations (\cref{sec:baselines}). Four paired repetitions compare generic and task-informed calibration within each configuration across six dense models (1.7--8 billion parameters), six data-dependent weight-only quantizers, and ten tasks.

At their lowest tested functional bit widths, iterative SignRound~V2 (SR-V2) and OmniQuant improve at two bits, whereas single-pass GPTQ and AWQ show small mean responses at three bits (\cref{tab:sensitivity}). At two bits, SR-V2 gains $1.83$ percentage points (pp) in mean ten-task accuracy across six models over Pile-200, a generic-text control with 200 requested calibration samples. At three bits, SR-V2 shows greater gap closure than AWQ with method-specific controls. Comparing accuracy gains with a common generic control does not establish a general class advantage (\cref{sec:bitwidth}). On Llama-3.1-70B (70 billion parameters), the two-bit GPTQ configuration is functional and gains $+2.33$ pp (\cref{sec:scale_verification}).

In the five-model two-bit SR-V2 controls, gains concentrate on families included in calibration. Marking wrong answers as correct retains about 90\% of the mean gain obtained with correct designations. Smaller mean gains remain when no answer is designated (\cref{sec:fragile_robust_tasks,sec:wrong_answer}).

SR-V2's data-dependent control without iterative updates shows little response. FragCal-DC has higher mean accuracy than the tested unranked question-answer mixture and other recipes (\cref{sec:recipe_controls}). Changes in reconstruction error, the difference between quantized and unquantized block outputs, do not consistently track accuracy gains.

We contribute cross-method calibration evidence at functional bit widths, task-level recovery controls, and the documented FragCal-DC recipe. Calibration mixtures should therefore be designed around the model's intended downstream tasks.

\section{Background and related work}
\label{sec:background}

\paragraph{Block-wise quantization.}
We study data-dependent block-wise PTQ, which fits low-bit weights to outputs of layers or Transformer blocks~\citep{frantar2023gptq,shao2024omniquant}. For calibration text $D$, let $f_{\ell}$ denote unit $\ell$'s computation, $\mathbf{X}_{\ell}(D)$ its input activations, and $\mathbf{W}_{\ell},\widehat{\mathbf{W}}_{\ell}$ its unquantized and quantized weights. A representative loss sums squared output differences (the squared Frobenius norm),
\begin{equation}
\mathcal{L}_{\ell}(\widehat{\mathbf{W}}_{\ell};D)=
\left\|f_{\ell}(\mathbf{W}_{\ell},\mathbf{X}_{\ell}(D))
-f_{\ell}(\widehat{\mathbf{W}}_{\ell},\mathbf{X}_{\ell}(D))\right\|_F^2.
\label{eq:block_loss}
\end{equation}
Calibration composition therefore determines the activations used to fit the low-bit weights. The reconstruction unit and objective vary by method.

\paragraph{Representations and updates.}
Methods differ in their low-bit representations and update rules. Within a reconstruction unit, each scalar-affine weight group uses scale $\gamma>0$, integer offset $z$, and integer codes $\mathbf{c}$ to represent weights as $\widehat{\mathbf{w}}=\gamma(\mathbf{c}-z)$. GPTQ uses second-order corrections~\citep{frantar2023gptq}, while AWQ~\citep{lin2024awq} and the zero-iteration RTN-opt control search scales. These single-pass methods omit repeated gradient-based reconstruction updates. AdaRound introduced local learned rounding through continuous rounding approximations~\citep{nagel2020adaround}. Iterative updates optimize OmniQuant's clipping thresholds~\citep{shao2024omniquant} or SR-V2's rounding and scales~\citep{cheng2025signroundv2}. AQLM iteratively learns vector codewords whose sums represent weights~\citep{egiazarian2024aqlm}.

\paragraph{Calibration composition.}
Prior studies report calibration effects that vary with the metric and setting. AWQ reports smaller cross-domain perplexity changes than GPTQ for three-bit OPT-6.7B~\citep[Fig.~8]{lin2024awq}. OmniQuant reports low perplexity variance across WikiText2, C4, and Pile for LLaMA-7B at three-bit weights with 16-bit activations and at four-bit weights and activations~\citep[Table~A10]{shao2024omniquant}. Across five sources and repeated draws, \citet{williams2024calibration} find task- and model-dependent accuracy variation, with lower sensitivity for four-bit GPTQ/SpQR than pruning and often small mean-accuracy differences between sources. These findings leave task-informed recovery below four bits unresolved.

\paragraph{Reconstruction and task accuracy.}
Reconstruction error measures local output fidelity. Holding the reconstruction metric and reference text fixed makes errors comparable across calibration conditions. Downstream task accuracy must also be measured because the local objective does not directly optimize it.

\paragraph{Scope.}
Our primary setting is weight quantization. We exclude QuIP/QuIP\# transforms~\citep{chee2024quip,tseng2024quipsharp}, sparse outlier representations (SpQR/SqueezeLLM~\citep{dettmers2024spqr,kim2024squeezellm}), and data-free HQQ~\citep{badri2023halfquadratic}. Our two-model weight-activation extension does not compare weight-activation algorithms or quantization-aware training~\citep{xiao2023smoothquant,li2023llmqat}. Floating-point quantization formats, key-value cache compression, mixed-precision allocation, and pruning are outside scope. We vary calibration composition while keeping update rules fixed (\cref{sec:setup}).

\section{Experimental setup}
\label{sec:setup}

\paragraph{Models and evaluation.}
The primary dense panel contains Qwen3-1.7B and Qwen3-8B~\citep{qwen3_2025}, Llama-2-7B~\citep{touvron2023llama2}, Mistral-7B-v0.1~\citep{jiang2023mistral}, Qwen2.5-7B-Instruct~\citep{qwen2p5_2024}, and Phi-3-mini-4k-instruct~\citep{abdin2024phi3}. Each model's unquantized full-precision (FP) accuracy serves as its reference. Llama-3.1-70B~\citep{dubey2024llama3} supplies a scale comparison.

The ten-task panel uses lm-eval-harness~\citep{gao2024lmeval} to evaluate ARC-Challenge and ARC-Easy~\citep{clark2018arc}, BoolQ~\citep{clark2019boolq}, HellaSwag~\citep{zellers2019hellaswag}, LAMBADA~\citep{paperno2016lambada}, MMLU~\citep{hendrycks2021mmlu}, OpenBookQA~\citep{mihaylov2018openbookqa}, PIQA~\citep{bisk2020piqa}, TruthfulQA MC1~\citep{lin2022truthfulqa}, and WinoGrande~\citep{sakaguchi2021winogrande}. AVG is their unweighted mean. Comparisons use matched evaluation settings within each method (\cref{app:srv2_repro}).

\paragraph{Quantization configurations.}
We test two-, three-, and four-bit weights (W2, W3, W4), with method coverage shown in \cref{tab:methods}. In calibration-response experiments, scalar weight quantizers use group size 128, denoted G128; bit width and group size combine as W2G128, W3G128, and W4G128. AQLM uses approximately two bits per weight before overhead. Sequence lengths and other implementation settings differ by method (\cref{app:hardware}).

\begin{table}[t]
\centering\small
\caption{Evaluated methods. \spclass~denotes single-pass and \itclass~iterative methods. Single-pass excludes repeated gradient-based reconstruction updates, not data dependence or search. RTN-opt uses SR-V2's scale-search initialization without iterative rounding updates.}
\label{tab:methods}
\begin{tabularx}{\linewidth}{@{}llX@{}}
\toprule
Method & Class & Evaluated implementation\\
\midrule
RTN & Data-free reference & Round-to-nearest\\
\spclass~RTN-opt & Single-pass & Importance-weighted scale search, zero iterations\\
\spclass~AWQ & Single-pass & Activation-aware scale search\\
\spclass~GPTQ & Single-pass & Hessian-based rounding corrections\\
\midrule
\itclass~OmniQuant & Iterative & Learnable weight clipping (LWC), 20 epochs\\
\itclass~SignRound V2 & Iterative & Rounding and scale optimization, 200 iterations\\
\itclass~AQLM & Iterative & Learned vector codebook, W2-equivalent\\
\bottomrule
\end{tabularx}
\end{table}

Implementation choices limit comparisons. Our OmniQuant uses learnable weight clipping (LWC) and disables learnable equivalent transformations (LET). These results do not characterize the full pipeline or establish a lower bound on its attainable accuracy. RTN-opt is a data-dependent zero-iteration control, but does not isolate iteration from other optimizer and parameterization choices.

\paragraph{Calibration intervention.}
FragCal-DC supplements generic text with training documents from fragile task families. An exploratory five-model SR-V2 scan with Pile calibration at two bits, group size 16, and seed 42 selects families with low ratios of quantized to unquantized accuracy (\cref{app:fragcal}). The 208-document mixture contains 128 Pile, 32 MMLU auxiliary-training, 32 from a combined ARC-Challenge/ARC-Easy/OpenBookQA training pool, and 16 HellaSwag training documents. Short question-answer items are concatenated into long documents before sampling.

Pile-128 and Pile-200 denote generic-text controls from Pile-10k~\citep{gao2020pile} with 128 and 200 requested calibration samples. Effective sequence counts depend on the tokenizer and loader (\cref{app:calibration_accounting}). Pile-200 is a larger generic control, not an exactly matched document or token budget. Calibration uses training splits, but held-out evaluation splits do not establish item-level separation (\cref{sec:overlap}). Paired responses use four calibration seeds, $\{42,123,456,789\}$, with model, quantizer configuration, and evaluation settings fixed. Single-seed scans and probes are labeled.

\paragraph{Comparison map.}
We distinguish the common-W2 observation (\cref{fig:method_type}), the designated-width comparison (\cref{tab:sensitivity}), and the matched-W3 check (\cref{sec:bitwidth}). The designated-width comparison and the matched-W3 check report Pile-200 pp effects across six models. The OmniQuant W2 comparison uses five models with functional baselines, excluding Mistral-7B. The task-level and paired answer-control analyses use five models with SR-V2/RTN-opt at W2 and Pile-200. Separately reported Phi-3 answer-control results are retained in the six-model descriptive summaries (\cref{tab:permodel_contam}).

\paragraph{Metrics and uncertainty.}
We report accuracy gains in percentage points (pp) and gap closure (GC). With $A$ denoting accuracy, per-model GC is $100(A_{\rm FragCal}-A_{\rm ctrl})/(A_{\rm FP}-A_{\rm ctrl})$ percent. For \cref{fig:method_type}, we average per-model GC. For \cref{tab:sensitivity,tab:bitwidth}, we apply the formula to mean accuracies over the same models as the pp gain. GC normalizes recovery by the remaining FP gap.

We average four paired-seed accuracy differences (FragCal minus control) per model, then across models. Panel means use 95\% Student $t$ intervals across models. Principal contrasts use exact two-sided model-level sign-flip tests, which need not agree with small-sample $t$ intervals. Seeds do not add independent models. Model independence is approximate because some models share architecture and training lineage. For \cref{fig:method_type}, annotated $p$ values are from unadjusted, two-sided one-sample $t$ tests against zero mean gap closure.

\section{Locating the functional operating regimes}
\label{sec:baselines}

We select operating widths using accuracy under generic calibration. For each model and quantizer configuration, we average ten-task accuracy over four seeds. A model's configuration is functional above $0.35$ and collapsed at or below it (\cref{app:cliff_regime}). The cutoff uses absolute accuracy, not full-precision retention. It is descriptive for this task suite, not a test against a common chance probability.

Functional widths differ across methods. \Cref{tab:baselines} summarizes the scan by method and width. For GPTQ and AWQ, all six W2 model baselines fall below $0.35$, so we compare these methods at W3. All six SR-V2 W2 baselines and all six AQLM W2-equivalent baselines are functional.

\begin{table}[htbp]
\centering\small
\caption{Generic-calibration scan of six models.
AVG averages unweighted ten-task accuracy (fraction) over available models.
FP is unquantized. Values use seed 42 except SR-V2 W4 (\cref{app:baseline_scan}).
AQLM is W2-equivalent, not scalar W2G128.
Panel-mean bands are \xmark~low ($\le0.35$),
\cmark~functional ($0.35<\mathrm{AVG}\le0.62$),
and $\uparrow$~high ($>0.62$). The $0.62$ guide does not affect selection.}
\label{tab:baselines}
\begin{tabular}{@{}lrrrrrrrr@{}}
\toprule
Width & FP Ref & RTN & \shortstack{\spclass\\RTN-opt} & \shortstack{\spclass\\AWQ} & \shortstack{\spclass\\GPTQ} & \shortstack{\itclass\\OmniQuant} & \shortstack{\itclass\\SR-V2} & \shortstack{\itclass\\AQLM}\\
\midrule
\inputrows{generated/baseline_mean_rows.tex}
\bottomrule
\end{tabular}
\end{table}

OmniQuant and RTN-opt use different model-selection rules. OmniQuant's Mistral-7B W2 baseline is $0.320$, below the threshold, whereas its other five baselines are functional. We report both the resulting five-model subset and the all-six-model response. RTN-opt remains a separate zero-iteration control on all six models, without threshold filtering.

These selection rules do not match bit widths or full-precision retention across methods. Raising the cutoff to $0.40$ would leave only two OmniQuant W2 models, too few for a broad comparison (\cref{app:cliff_regime}). We therefore accompany the operating-width comparison with matched-W3 and retention-matching checks (\cref{sec:bitwidth}).

\section{Calibration sensitivity across methods and regimes}
\label{sec:results}

\subsection{Response at the designated operating widths}

The strongest mean responses at the designated widths come from SR-V2 on six models and OmniQuant on the five models with functional W2 baselines (\cref{tab:sensitivity}). Their W2 mean effects are $+1.83$ and $+1.78$ pp, respectively, with 6/6 and 5/5 positive model means. AQLM's six-model estimate is $+0.73$ pp, with five positive model means, but its interval includes zero. GPTQ and AWQ at W3, and RTN-opt at W2, have small panel means with intervals spanning zero (3/6, 4/6, and 2/6 positive models, respectively).

\begin{table}[htbp]
\centering\small
\caption{Calibration sensitivity at the designated operating widths.
AVG is unweighted ten-task accuracy (fraction),
averaged over four paired seeds and $n$ models.
$\Delta$ is FragCal-DC minus Pile-200 (pp), with model-level 95\% $t$ intervals.
GC (\%) is FP gap closure from the same mean accuracies.
OmniQuant W2 excludes Mistral-7B ($n=5$).}
\label{tab:sensitivity}
\begin{tabular}{@{}llcrrrrr@{}}
\toprule
Method & Format & $n$ & \shortstack{Pile-200\\AVG} &
\shortstack{FragCal-DC\\AVG} & $\Delta$ (pp) & 95\% CI & GC (\%)\\
\midrule
\inputrows{generated/method_rows.tex}
\bottomrule
\end{tabular}
\end{table}

OmniQuant's result depends on model selection and how uncertainty is assessed. Including the below-threshold Mistral-7B result gives an all-six-model W2 estimate of $+1.20$ pp with interval $[-0.34,+2.73]$. The exact two-sided sign-flip $p$ values for the pp effects are $0.03125$ for SR-V2 and $0.0625$ for OmniQuant's five-model subset. The $0.0625$ value is the smallest attainable with five models, whereas OmniQuant's positive Student $t$ interval uses a different sampling assumption.

\subsection{Matched-width comparison}
\label{sec:bitwidth}

At W3, SR-V2 shows greater per-model gap closure than AWQ with the method-specific control references (paired $t(5)=3.40$, one-sided $p=0.010$, \cref{app:matched}). Calibration sample counts are not matched.

Accuracy-gain comparisons with common Pile-200 controls do not establish a general W3 class advantage. SR-V2's accuracy gain is $+0.40$ pp higher than AWQ's, with 95\% interval $[-0.10,+0.90]$. Averaging SR-V2/OmniQuant versus GPTQ/AWQ gives a class contrast of $+0.06$ pp, with interval $[-0.68,+0.80]$ and exact sign-flip $p=0.875$. Three of six model contrasts are positive.

Mean W4 responses are small, with $+0.01$ pp for SR-V2, $-0.08$ for OmniQuant, $-0.07$ for GPTQ, and $-0.03$ for AWQ (\cref{tab:bitwidth}).

Baseline accuracy and distance to collapse vary even at the same bit width. Retention matching at the designated operating widths yields only one or two model pairs under the tested tolerances, all OmniQuant versus RTN-opt, and cannot support a broad class inference. The available configuration grid does not isolate an optimizer-class effect from baseline accuracy, codebook, objective, or initialization.

\begin{table}[htbp]
\centering\small
\caption{FragCal-DC accuracy gains ($\Delta$, pp) and FP gap closure (GC, \%) over Pile-200.
Both use six-model mean accuracies, excluding Mistral-7B for OmniQuant W2 ($n=5$).
AQLM uses a W2-equivalent codebook. Dashes mark inapplicable widths.
Per-model results are in \cref{tab:cross_models}.}
\label{tab:bitwidth}
\begin{tabular}{@{}llrrrrrr@{}}
\toprule
Method & Type & \multicolumn{2}{c}{W2G128} & \multicolumn{2}{c}{W3G128} & \multicolumn{2}{c}{W4G128}\\
\cmidrule(lr){3-4}\cmidrule(lr){5-6}\cmidrule(l){7-8}
& & $\Delta$ (pp) & GC (\%) & $\Delta$ (pp) & GC (\%) & $\Delta$ (pp) & GC (\%)\\
\midrule
\inputrows{generated/width_rows.tex}
\bottomrule
\end{tabular}
\end{table}

\subsection{Scale and architecture boundary cases}
\label{sec:scale_verification}

Both GPTQ and SR-V2 improve Llama-3.1-70B at W2, where their generic-calibration baselines are functional. Over four paired seeds, GPTQ gains $+2.33$ pp (approximately $0.367$ to $0.390$ AVG), while SR-V2 gains $+0.86$ pp ($0.659$ to $0.668$ AVG, \cref{tab:seventy_b,app:70b_detail}). Gap closure is $6.6\%$ for GPTQ and $13.8\%$ for SR-V2, reversing their ranking by absolute gain. These results extend calibration sensitivity across scale and limit how broadly the smaller-model method-class pattern applies.

\begin{table}[htbp]
\centering\small
\caption{Llama-3.1-70B W2G128 calibration comparison.
Accuracy is the ten-task mean (fraction). Rows average four seeds except AWQ (seed 42).
The unquantized bfloat16 (BF16, 16-bit floating-point) reference has mean accuracy 0.7215.
$\Delta$ is FragCal minus Pile (pp).
GC is the percentage of the gap between Pile and BF16 accuracy recovered by FragCal.}
\label{tab:seventy_b}
\begin{tabular}{@{}lcrrrr@{}}
\toprule
Method & Seeds & Pile-200 & FragCal & $\Delta$ pp & GC (\%)\\
\midrule
\spclass~RTN-opt & 4 & .3213 & .3239 & +0.26 & +0.6\\
\spclass~AWQ & 1 & .322 & .321 & $-0.1$ & $-0.3$\\
\spclass~GPTQ & 4 & .3665 & .3899 & +2.33 & +6.6\\
\midrule
\itclass~SR-V2 & 4 & .659 & .668 & +0.86 & +13.8\\
\bottomrule
\end{tabular}
\end{table}

Positive iterative responses also appear in narrower architecture and format tests (\cref{app:extensions}). On Mixtral-8x7B-Instruct, a mixture-of-experts model that routes inputs to selected expert networks, SR-V2 improves by $+1.02$ pp and GPTQ changes by $-1.39$ pp at W3G128 on ARC-Challenge only, over four seeds. AutoRound also improves Phi-3 and Mistral with two-bit weights and eight-bit activations under both tested formats. These are not general architecture or weight-activation class comparisons, and routing changes were not measured.

\subsection{Where the accuracy gain occurs}
\label{sec:fragile_robust_tasks}

SR-V2's W2 recovery is concentrated in the three fragility-selected tasks. MMLU, ARC-Challenge, and HellaSwag gain $+3.86$ pp on the five-model Pile-200 panel, compared with $+0.20$ pp on five strict non-targeted tasks (\cref{tab:targeting}). The strict group excludes the selected tasks plus ARC-Easy and OpenBookQA, whose training data also enter calibration. The paired within-model difference is $+3.65$ pp. This supports task-aligned recovery but does not separate fragility from calibration-source exposure.

\begin{table}[htbp]
\centering\small
\caption{SR-V2 W2G128 FragCal gains over Pile-200 (pp),
averaged over four paired seeds and five models, with model-level 95\% $t$ intervals.
Fragility-selected tasks are MMLU, ARC-Challenge, and HellaSwag.
Strict non-targeted tasks exclude these plus ARC-Easy and OpenBookQA.
The last row is the within-model difference between selected and strict non-targeted gains.}
\label{tab:targeting}
\begin{tabular}{@{}lrrc@{}}
\toprule
Task group & $\Delta$ pp & 95\% interval & Positive models\\
\midrule
\inputrows{generated/targeting_rows.tex}
\bottomrule
\end{tabular}
\end{table}

\begin{figure}[htbp]
\centering
\includegraphics[width=\linewidth]{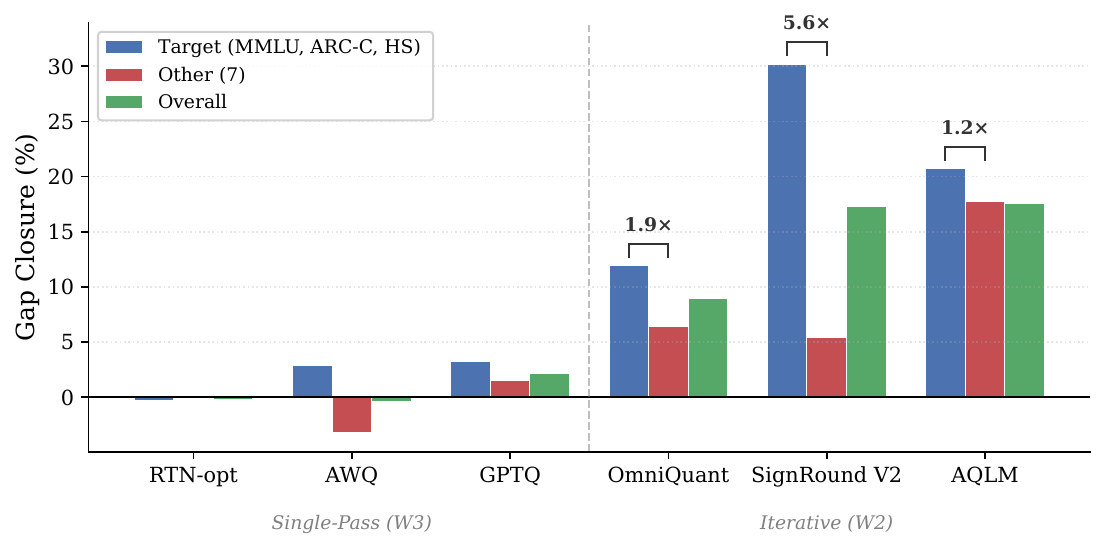}
\caption{Task-group gap closure is the percentage of the gap between FP and Pile accuracy recovered by FragCal.
Bars show MMLU, ARC-Challenge, and HellaSwag (blue), the other seven tasks (red), and all ten (green).
Brackets give blue/red ratios.
Red bars include ARC-Easy and OpenBookQA, unlike the strict controls
in \cref{tab:targeting}. This descriptive panel uses W3 single-pass and W2
iterative methods (AQLM W2-equivalent), with SR-V2's Pile-128 reference.}
\label{fig:fragile_robust}
\end{figure}

Task-group gap closure provides a descriptive view of recovery across quantizers (\cref{fig:fragile_robust}). Unlike \cref{tab:targeting}, it uses mixed operating widths and SR-V2's Pile-128 reference. Its seven-task complement includes ARC-Easy and OpenBookQA, so it is not an independent-transfer control.

\section{Calibration controls and explanatory evidence}
\label{sec:contamination}

The task concentration in \cref{sec:fragile_robust_tasks} raises questions
about answer content, recipe choice, transfer, and overlap. We examine these
before testing reconstruction and scale-code explanations.

\subsection{Answer designation and zero-iteration controls}
\label{sec:wrong_answer}

Wrong answer designations preserve most of the SR-V2 W2 improvement, while omitting designation preserves a smaller mean gain (\cref{tab:wrong_answer}). On the five-model Pile-200 panel, correct, incorrect, and omitted designations yield $+1.60$, $+1.44$, and $+0.88$ pp, respectively. Wrong designations retain about 90\% of the correct-answer mean gain, and no designation retains about 55\%. The separate six-model comparison with method-specific references retains approximately 91\% under wrong designations (\cref{tab:permodel_contam}).

These interventions test answer designation rather than removal of all answer information. MMLU, ARC, and OpenBookQA retain all candidate options, including the correct one.

\begin{table}[htbp]
\centering\small
\caption{Answer-designation and zero-iteration controls, W2G128, five models,
four paired seeds. AVG is mean ten-task accuracy (fraction).
$\Delta$ is the gain over each method's Pile-200 control (pp), with
model-level 95\% $t$ intervals. Gain retained is relative to the
correct-answer SR-V2 mean gain. No designation leaves answers unmarked.}
\label{tab:wrong_answer}
\begin{tabular}{@{}lrrrr@{}}
\toprule
Calibration & AVG & $\Delta$ pp & 95\% interval & Gain retained\\
\midrule
\inputrows{generated/answer_summary_rows.tex}
\bottomrule
\end{tabular}
\end{table}

The data-dependent zero-iteration RTN-opt control shows little mean response on the same five models, at $-0.04$ pp with interval $[-0.35,+0.28]$. This supports iterative optimization as a plausible contributor in the tested framework, without isolating its effect. Iteration-count and optimizer-variant probes provide complementary evidence (\cref{app:iteration_ablation,app:dm_signsgd}).

\subsection{Does the recipe matter?}
\label{sec:recipe_controls}

Random QA captures much of FragCal's mean benefit, but with more regressions. This unranked question-answer mixture gains $+1.46$ pp versus $+1.93$ pp for FragCal on the six-model SR-V2 W2 panel (\cref{app:random_qa}). References are labeled Pile-128 for five models and Pile-200 for Phi-3. Random QA has 18/24 positive model-seed pairs, versus 24/24 for FragCal. Its 200 question-answer items without a Pile anchor differ from FragCal's 208 documents in both source allocation and generic-text coverage, so the comparison does not isolate fragility ranking.

FragCal has higher mean accuracy than three tested alternatives based on diversity selection, semantic selection, and scientific prose. On the five-model SR-V2 W2 panel, same-harness advantages are $+2.01$ pp over Diversity-DC, $+1.11$ over LSA-DC, and $+2.55$ over ArXiv-OOD (\cref{app:alternatives}). The LSA-DC interval includes zero, and these comparisons do not establish an optimal ranking.

\subsection{Transfer and evaluation overlap}
\label{sec:overlap}

ARC/OpenBookQA supplements yield larger MMLU gains than MMLU supplements in the tested comparison. Both arms add 32 documents to Pile-128 and are evaluated on MMLU with SR-V2 W2. The ARC/OpenBookQA advantage is $+1.50$ pp, with hierarchical-bootstrap interval $[+0.67,+2.46]$, positive in all five model means and 19/20 seed pairs (\cref{app:transfer}). Direct MMLU supplementation is therefore not needed for the stronger MMLU response in this comparison. Both sources contain question-answer text, leaving activation coverage, format, and distribution alignment as possible explanations.

The more distant-task evaluations do not establish broad cross-family transfer (\cref{app:gsm8k,app:orthogonal}). They cover SR-V2 and RTN-opt on GSM8K, and GPTQ, AWQ, OmniQuant, and AQLM on GPQA-Diamond, IFEval, and BBH. Their differing method coverage prevents treating them as a single cross-method replication.

\paragraph{Item-level overlap.}
The overlap check finds matches for normalized evaluation questions and HellaSwag completions in the calibration files (\cref{tab:overlap}). Some question keys are generic, and the question matches have not been verified as full question-plus-options duplicates. Named split separation does not establish item separation. Full-item verification and the accuracy impact of removing matched items remain unresolved. Neither wrong answers nor the external-task measurements establish contamination-free gains.

\subsection{Reconstruction proxies and possible mechanisms}
\label{sec:loss_paradox}

\paragraph{Does reconstruction error track task recovery?}
Reconstruction comparisons require common inputs. Losses on different calibration sets are not directly comparable, and method-specific objectives are not interchangeable. We use the same 128 Pile-10k samples at seed 999 (\cref{app:calibration_loss}). Accuracy uses four-seed means. Relative reconstruction-error changes are averaged over available paired seeds, whose coverage varies (\cref{tab:heldout_permodel}).

On this common reference, reconstruction error and accuracy do not consistently improve together (\cref{tab:loss_paradox}). GPTQ's error increases in five of six model cells, yet accuracy improves in three. AQLM's error increases in all six cells, yet accuracy improves in five.

\begin{table}[htbp]
\centering\small
\caption{Reconstruction error and accuracy changes from Pile-200 to FragCal, averaged over $n$ models.
MSE denotes fixed-reference mean squared reconstruction error.
Lower MSE and higher accuracy are better.
Mismatch counts models where one metric improves and the other worsens.}
\label{tab:loss_paradox}
\begin{tabular}{@{}lcrrr@{}}
\toprule
Method & $n$ & $\Delta$ MSE (\%) & $\Delta$ accuracy (pp) & Mismatch\\
\midrule
\inputrows{generated/loss_summary_rows.tex}
\bottomrule
\end{tabular}
\end{table}

Two optimization probes also show that lower loss need not improve accuracy. The optimizer variant reduces calibration loss by 6--30\%. A second pass lowers its own objective but worsens 15/20 accuracy cells (\cref{app:dm_signsgd}). That pass starts from already quantized weights, changing the reconstruction problem.

\paragraph{Can another loss predict task recovery?}
Held-out next-token prediction loss (cross-entropy) shows a preliminary association with task recovery. Its reduction on WikiText correlates with accuracy gain across five SR-V2 W2 model means (Pearson $r=0.778$, exact permutation $p=0.083$, \cref{app:wikitext_proxy}). The interval for the mean cross-entropy reduction includes zero. This small panel does not validate a predictor of recovery.

\paragraph{Do scale changes explain calibration sensitivity?}\label{sec:mechanism}
Group-scale spread is more sensitive to calibration composition for SR-V2 than GPTQ in the nine tested model-width cells (\cref{tab:mechanism_3family,app:mechanism}). We measure spread with the coefficient of variation (CV), the scale standard deviation divided by the magnitude of the mean. The SR-V2/GPTQ sensitivity ratio is largest at W4 for two models despite small accuracy gains. This ratio is therefore not a standalone accuracy predictor.

\begin{table}[htbp]
\centering\small
\caption{Scale sensitivity to replacing Pile with FragCal (SR-V2/GPTQ ratios).
For per-group scales $s$, scale CV is
$\sigma(s)/(|\mu(s)|+10^{-12})$, where $\sigma$ and $\mu$ are their population
standard deviation and mean. Sensitivity is the relative absolute CV change, averaged over
matched modules and then four seeds (\cref{app:mechanism}).
Phi-3 uses SR-V2's signed-scale convention.}
\label{tab:mechanism_3family}
\begin{tabular}{@{}lrrr@{}}
\toprule
Model & W2G128 & W3G128 & W4G128\\
\midrule
Llama-2-7B & $16.4\times$ & $68.1\times$ & $164.5\times$\\
Qwen3-1.7B & $15.8\times$ & $24.8\times$ & $41.5\times$\\
Phi-3 & $1.2\times$ & $4.8\times$ & $4.2\times$\\
\bottomrule
\end{tabular}
\end{table}

Flattening learned scales while holding codes fixed harms fidelity, supporting the importance of the learned scale-code pairing. Whether reoptimized codes could compensate for flatter scales remains untested. The intervention does not establish scale variation as a sufficient cause of FragCal's accuracy gains.

Calibration also changes GPTQ's input statistics, but the probe does not identify which rounding decisions change (\cref{app:hessian}). Calibration-dependent optimization and distribution alignment remain compatible explanations.

\section{Discussion}
\label{sec:discussion}

Calibration design begins with the quantized configuration. The baseline scan identifies where each method retains useful accuracy (\cref{sec:baselines}), and the response comparisons measure the effect of task-informed data at those widths (\cref{sec:results}). Variation across widths and the positive GPTQ response at 70B motivate evaluating recipes for the intended model, quantizer, and bit width.

Within that configuration, task coverage guides calibration composition. The two-bit SR-V2 controls support task-aligned recovery. Gains concentrate on families included in calibration and largely persist with incorrect answer designations (\cref{sec:fragile_robust_tasks,sec:wrong_answer}). This makes task content and coverage useful starting points for recipe design. FragCal-DC implements this approach through a mixture of task-relevant training examples and generic text.

The reconstruction and scale probes in \cref{sec:loss_paradox} reinforce the need to judge these choices by downstream accuracy. Reconstruction objectives measure local approximation quality, whereas task accuracy reflects decisions made by the complete model. Reconstruction and scale measurements help characterize quantization behavior, while task accuracy determines whether a calibration change is useful.

\paragraph{Practical use.}\label{sec:protocol}
Start from a generic-calibration baseline with useful task accuracy. Compare mixtures of suitable training data and generic text while holding other quantization settings fixed. Evaluate target tasks and independent controls, separate recipe selection from final testing, and check item-level overlap.

\paragraph{Limitations.}\label{sec:limitations}
Optimizer class remains coupled to width, baseline accuracy, parameterization, and implementation. The small panel includes related architectures, and broad transfer remains unestablished. Code generation and multi-turn dialogue are untested. FragCal-DC's ranking and allocation need independent validation across seeds and new target distributions. The accuracy effect of overlapping calibration and evaluation items remains unresolved.

\section{Conclusion}
\label{sec:conclusion}

Calibration composition can recover downstream accuracy while leaving the quantizer's update rule unchanged. At the lowest tested functional widths, recovery is clearest for two-bit SignRound~V2 and OmniQuant's five-model subset, with small mean responses for three-bit GPTQ and AWQ. On Llama-3.1-70B, both GPTQ and SignRound~V2 improve at two bits. For two-bit SignRound~V2, gains concentrate on task families included in calibration and largely persist with incorrect answer designations. FragCal-DC is one recipe for task-informed calibration. These findings support task-informed calibration as a design principle for preserving downstream accuracy in low-bit quantization.

\paragraph{Societal considerations.}
Efficient inference can reduce deployment costs and resource use, but can also facilitate harmful uses of existing capabilities. Calibration may selectively preserve or degrade capabilities, so aggregate accuracy alone is insufficient for safe deployment. This work involves no human-subject study. Use of the calibration protocol must respect the source datasets' and models' terms.

\section*{Acknowledgments and disclosures}
This work received no third-party or agency funding.
Beyond the affiliations listed above, the authors declare no additional competing interests.

\bibliographystyle{plainnat}
\bibliography{references}


\appendix
\crefalias{section}{appendix}
\crefalias{subsection}{appendix}
\crefalias{subsubsection}{appendix}
\noindent\textbf{Appendix guide.}
Per-model results and regime checks are in
\cref{app:per_model,app:matched,app:cliff_regime}; calibration construction
and implementation details are in \cref{app:fragcal,app:fragcal_script,app:hardware}.
Additional reconstruction and scale-code evidence is in
\cref{app:calibration_loss,app:dm_signsgd,app:mechanism};
recipe, transfer, overlap, and architecture controls are in \cref{app:rebuttal}.

\section{Per-model calibration responses}
\label{app:per_model}

These tables provide per-model evidence for
\cref{sec:baselines,sec:results,sec:wrong_answer}. Four-seed responses appear
first, followed by the baseline scan, FP references, and answer-designation
controls. Each row of \cref{tab:primary_models} averages four seeds.
The task-group intervals in \cref{tab:targeting} use these five model means.

\begin{table}[H]
\centering\small
\caption{Five-model W2G128 accuracy gains with FragCal over Pile-200 (pp), averaged over four seeds.
AVG averages ten tasks. Task groups are defined in
\cref{tab:targeting}.}
\label{tab:primary_models}
\begin{tabular}{@{}lrrrr@{}}
\toprule
Model & \spclass~RTN-opt & \multicolumn{3}{c}{\itclass~SR-V2}\\
\cmidrule(lr){3-5}
& AVG & AVG & Fragility-selected & Strict non-targeted\\
\midrule
\inputrows{generated/primary_model_rows.tex}
\bottomrule
\end{tabular}
\end{table}

The cross-method comparisons use Pile-200 throughout, including Phi-3 for
SR-V2. OmniQuant Mistral-7B at W2 remains in \cref{tab:cross_models} but is
excluded from the means in \cref{tab:sensitivity,tab:bitwidth} because its
Pile baseline is below 0.35.

\clearpage
{\small\setlength{\tabcolsep}{3pt}
\begin{longtable}{@{}lll lrrrr@{}}
\caption{Four-seed ten-task means for all 96 cross-method model records.
Pile and FragCal are accuracies (fractions).
$\Delta$ is FragCal minus Pile (pp). GC (\%) uses unrounded per-model FP
references. W2-eq.\ denotes AQLM's vector codebook.}
\label{tab:cross_models}\\
\toprule
Method & Model & Format & Reference & Pile & FragCal & $\Delta$ pp & GC\\
\midrule\endfirsthead
\multicolumn{8}{@{}l}{Table \thetable\ (continued)}\\
\toprule
Method & Model & Format & Reference & Pile & FragCal & $\Delta$ pp & GC\\
\midrule\endhead
\bottomrule\endfoot
\inputrows{generated/cross_model_rows.tex}
\end{longtable}
}

GC magnitudes can become large when the baseline is close to FP accuracy.
Interpret them alongside the absolute accuracy change in pp.

\paragraph{Calibration-loader accounting.}
\label{app:calibration_accounting}
Sample-count labels describe requested calibration sizes. Effective volumes
vary by implementation. AWQ and OmniQuant use 512-token calibration blocks.
\begin{itemize}[itemsep=1pt]
\item \textbf{AWQ.} At W4, 19 Pile controls contain 90--104 blocks and five
contain 200 blocks. All FragCal runs contain 208 blocks. The five 200-block
controls are the four Phi-3 seeds and Mistral seed 456. The latter uses
98 blocks at W2/W3 and 200 at W4.
\item \textbf{OmniQuant.} Controls contain 200 blocks and FragCal runs 208.
\item \textbf{RTN-opt.} The 20 W4 control configurations outside Phi-3 do not
record calibration sample counts.
\item \textbf{AQLM.} Both conditions request 200 samples for Llama-2-7B,
Mistral-7B, and Phi-3. The three Qwen models request 200 Pile and
208 FragCal samples. Its source files still contain 208 FragCal documents.
\end{itemize}
All recorded pairs are retained. The varying calibration volumes prevent
interpreting these comparisons as matched-token-budget or isolated bit-width ablations.

\subsection{Single-seed baseline scan}
\label{app:baseline_scan}

The tables below report seed-42 generic-Pile scan measurements, including
RTN and W4 coverage. The Pile-10k source-corpus name does not specify
the number of documents consumed by each implementation. Threshold selection
and the main response estimates use the four-seed paired panel reported above.
Means in \cref{tab:baselines} use the available scan rows, except for SR-V2
W4: its separately reported mean is $0.633$, without individual scan rows.

\begin{table}[H]
\centering\small
\caption{W2G128 seed-42 baseline scan, unweighted ten-task mean accuracy
(AVG, fraction). FP Ref is unquantized. AQLM is W2-equivalent,
not a scalar W2G128 quantizer.}
\label{tab:scan_w2}
\begin{tabular}{@{}lrrrrrrrr@{}}
\toprule
Model & FP Ref & RTN & \shortstack{\spclass\\RTN-opt} & \shortstack{\spclass\\AWQ} & \shortstack{\spclass\\GPTQ} & \shortstack{\itclass\\OmniQuant} & \shortstack{\itclass\\SR-V2} & \shortstack{\itclass\\AQLM}\\
\midrule
\inputrows{generated/baseline_w2_rows.tex}
\bottomrule
\end{tabular}
\end{table}

\begin{table}[H]
\centering\small
\caption{W3G128 seed-42 baseline scan, unweighted ten-task mean accuracy
(AVG, fraction). FP Ref is unquantized.}
\label{tab:scan_w3}
\begin{tabular}{@{}lrrrrrrr@{}}
\toprule
Model & FP Ref & RTN & \shortstack{\spclass\\RTN-opt} & \shortstack{\spclass\\AWQ} & \shortstack{\spclass\\GPTQ} & \shortstack{\itclass\\OmniQuant} & \shortstack{\itclass\\SR-V2}\\
\midrule
\inputrows{generated/baseline_w3_rows.tex}
\bottomrule
\end{tabular}
\end{table}

\begin{table}[H]
\centering\small
\caption{W4G128 seed-42 baseline scan, ten-task mean accuracy (fraction).
GPTQ's Llama-2 entry is omitted for an anomalous TruthfulQA score of 1.0.
No replacement is imputed.}
\label{tab:scan_w4}
\begin{tabular}{@{}lrrrrrr@{}}
\toprule
Model & FP Ref & RTN & \shortstack{\spclass\\RTN-opt} & \shortstack{\spclass\\AWQ} & \shortstack{\spclass\\GPTQ} & \shortstack{\itclass\\OmniQuant}\\
\midrule
\inputrows{generated/baseline_w4_rows.tex}
\bottomrule
\end{tabular}
\end{table}

\subsection{Full-precision task detail}

\begin{table}[H]
\centering\small
\caption{Width-scan FP per-task accuracies (fractions).
AVG is the unweighted ten-task mean and TruthfulQA uses MC1.
Same-harness rescoring uses separate references.
Only Phi-3's log records BF16. The other five loading dtypes are unrecorded.}
\label{tab:permodel_fp}
\begin{tabular}{@{}lcccccc@{}}
\toprule
Task & \shortstack{Qwen3\\1.7B} & \shortstack{Llama-2\\7B} &
\shortstack{Mistral\\7B} & \shortstack{Qwen2.5\\7B-Instruct} &
\shortstack{Qwen3\\8B} & \shortstack{Phi-3\\mini}\\
\midrule
\inputrows{generated/fp_task_rows.tex}
\bottomrule
\end{tabular}
\end{table}

\subsection{Answer-designation controls}

\begin{table}[H]
\centering\small
\caption{SR-V2 W2G128 ten-task accuracy gains with FragCal over Pile-200 (pp).
The first five rows use four paired seeds. Phi-3 is reported separately and excluded
from the intervals in \cref{tab:wrong_answer}.
The descriptive six-model mean uses the ratio of panel-mean gains in its last column.}
\label{tab:permodel_contam}
\begin{tabular}{@{}lrrrr@{}}
\toprule
Model & Correct & No designation & Wrong & Wrong/correct\\
\midrule
\inputrows{generated/answer_model_rows.tex}
\bottomrule
\end{tabular}
\end{table}

With the separate method-specific references (Pile-128 for the first five
models and Pile-200 for Phi-3), the six-model gains are $+1.93$ pp with
correct designations, $+1.75$ with wrong designations, and $+1.19$ with no answer
designated. Wrong and omitted designations retain approximately 91\% and 62\%
of the correct-answer gain in that comparison. These rounded descriptive
summaries use different controls from the Pile-200 means above and a
different model population from the five-model intervals.

\section{Matched-width comparison}
\label{app:matched}

At W3, we compare SR-V2 and AWQ by per-model gap closure and assess the
broader method-class contrast using accuracy gains over Pile-200.
These analyses address the matched-width question in \cref{sec:bitwidth}.

\paragraph{Normalized pairwise recovery.}
For each model and method, we average four calibration seeds within each arm.
The comparison covers six models.
Writing the FragCal, Pile, and FP mean accuracy fractions as $F_{m,j}$,
$P_{m,j}$, and $R_m$, per-model gap closure is
$g_{m,j}=100(F_{m,j}-P_{m,j})/(R_m-P_{m,j})$.
Pairing SR-V2 and AWQ at W3 gives $t(5)=3.40$, one-sided $p=0.010$
(two-sided $p=0.019$). AWQ uses Pile-200. SR-V2 uses the references
reported as Pile-128 except Phi-3, which uses Pile-200. The older SR-V2
configs for the other five models do not record sample counts.
This calculation uses the FP averages printed to three decimals in
\cref{tab:permodel_fp}. Unrounded FP averages give one-sided $p=0.0098$.
The paired test uses per-model GC differences.
\Cref{tab:sensitivity,tab:bitwidth} instead compute GC from panel-mean
accuracies. Control sample counts differ between the methods.

\paragraph{Common-reference accuracy gains.}
For each model $m$, let $d_{m,j}$ denote the W3 accuracy change for method
$j$ with FragCal relative to the control (pp). The class contrast is
\[
c_m=\tfrac12(d_{m,\mathrm{SRV2}}+d_{m,\mathrm{OQ}})
-\tfrac12(d_{m,\mathrm{GPTQ}}+d_{m,\mathrm{AWQ}}).
\]
All four methods use Pile-200 controls for the six models below. This is a width-matched
comparison, not a fully controlled optimizer-class intervention.
The separate SR-V2/AWQ accuracy-gain contrast is $+0.40$ pp, with model-level
95\% $t$ interval $[-0.10,+0.90]$ and exact two-sided sign-flip $p=0.09375$.

\begin{table}[H]
\centering\small
\caption{Within-model W3G128 contrasts of SR-V2/OmniQuant mean response
minus GPTQ/AWQ mean response (pp). The mean is $+0.06$ pp, 95\% model-level
$t$ interval $[-0.68,+0.80]$, exact two-sided sign-flip $p=0.875$.}
\label{tab:same_w3}
\begin{tabular}{@{}lr@{}}
\toprule
Model & Iterative minus single-pass (pp)\\
\midrule
\inputrows{generated/same_w3_rows.tex}
\bottomrule
\end{tabular}
\end{table}

The retention-matching audit finds only one or two paired models under the
tested tolerances, with every supported pair comparing OmniQuant and RTN-opt.
This sparse overlap does not identify a class-wide effect.
The common-reference class interval remains compatible with effects in either direction.

\section{Functional-regime identification}
\label{app:cliff_regime}

The operating rule in \cref{sec:baselines} uses the four-seed mean
generic-calibration ten-task AVG. A cell with AVG at or below 0.35 is
designated below the functional threshold. The cutoff measures absolute
accuracy on this task mixture. It is distinct from chance accuracy and
FP-relative retention, and does not predict calibration sensitivity.
The 0.62 upper guide describes higher-accuracy cells but does not enter
the exclusion rule.

\begin{table}[H]
\centering\small
\caption{Per-model four-seed ten-task mean accuracies (AVG, fractions).
A dagger marks
AVG at or below 0.35. All methods use Pile-200.
The W2 AQLM column is W2-equivalent.}
\label{tab:cliff_regime}
\begin{tabular}{@{}llrrrrr@{}}
\toprule
Width & Model & \shortstack{\spclass\\AWQ} & \shortstack{\spclass\\GPTQ} & \shortstack{\itclass\\OmniQuant} & \shortstack{\itclass\\SR-V2} & \shortstack{\itclass\\AQLM}\\
\midrule
\inputrows{generated/regime_rows.tex}
\bottomrule
\end{tabular}
\end{table}

The per-model values are listed with their references in
\cref{tab:cross_models}. All twelve GPTQ/AWQ W2 cells and the OmniQuant Mistral
W2 cell lie below 0.35. GPTQ and AWQ are therefore compared at W3, where all
six models are above the threshold. Only Mistral is excluded within
OmniQuant's designated W2 panel. All six SR-V2 W2 baselines and all six
AQLM W2-equivalent baselines exceed the threshold. Three AQLM baselines
exceed 0.62, so the functional group includes higher-accuracy configurations.

Including Mistral changes OmniQuant's model-level estimate from $+1.78$ pp
with 95\% interval $[+1.28,+2.27]$ to $+1.20$ pp with interval
$[-0.34,+2.73]$.

\paragraph{Sensitivity to the threshold.}
Raising the threshold to 0.40 also removes OmniQuant Llama-2-7B (0.3820),
Qwen3-1.7B (0.3658), and Phi-3 (0.3903), leaving Qwen2.5-7B-Instruct and
Qwen3-8B. Their mean response is approximately $+1.93$ pp, but two models
are insufficient for a stable generalization. The designated SR-V2, AQLM,
GPTQ, and AWQ panels do not change under this threshold adjustment.

RTN-opt is a separate zero-iteration control. Its six-model Pile-200 comparison
in \cref{tab:sensitivity} has no threshold filtering, including
Qwen3-1.7B's low baseline.

\section{FragCal-DC construction details}
\label{app:fragcal}

The recipe combines task-family selection with a supplement-size sweep.
The resulting fixed allocation is used in the response experiments
(\cref{sec:setup}).

\subsection{Exploratory fragility ranking (W2G16, seed=42)}

Task-family selection uses an exploratory SR-V2 W2G16 scan with Pile
calibration across five models:
Llama-2-7B, Mistral-7B, Qwen2.5-7B-Instruct, Qwen3-1.7B, and Qwen3-8B.
Phi-3 is not part of this selection panel. This G16 construction diagnostic
is distinct from the G128 dose-response and calibration-response experiments.
Full-precision references use automatic dtype selection for each model.
The single-seed ranking does not establish stability across seeds or new
model families.

\begin{table}[H]
\centering
\small
\caption{W2G16 construction diagnostic (five models, seed 42).
Retention (\%) averages within-model ratios of quantized to full-precision task accuracy, multiplied by 100.
Lower retention means greater fragility. Bold marks the three families prioritized for construction.}
\label{tab:fragility_ranking}
\begin{tabular}{@{}lr@{}}
\toprule
Task & FP retention \\
\midrule
\textbf{MMLU}          & 78.2\% \\
\textbf{ARC-Challenge} & 84.3\% \\
\textbf{HellaSwag}     & 86.0\% \\
OpenBookQA            & 89.7\% \\
LAMBADA               & 91.1\% \\
TruthfulQA            & 91.3\% \\
ARC-Easy              & 92.6\% \\
PIQA                  & 95.3\% \\
WinoGrande            & 95.8\% \\
BoolQ                 & 95.9\% \\
\bottomrule
\end{tabular}
\end{table}

\subsection{Dose-response summary}

The sweep tests how supplement size affects ten-task and per-task accuracy.
Both tables below use the same five-model sweep, evaluated with SR-V2 at
W2G128 and seed 42. Each supplement file appends $\{16,32,64\}$ training
documents to the same 128-document Pile control. Requested sample caps
match file sizes. The control yields 110--121 valid sequences after
tokenization and filtering, so document counts are not effective sequence counts.

\begin{table}[H]
\centering
\small
\caption{Five-model ten-task mean accuracy gains over the sweep's Pile-only control (pp).
Column headings count added raw documents. Best size maximizes the observed pooled gain.}
\label{tab:fragcal_dose_response}
\begin{tabular}{@{}lcccc@{}}
\toprule
Supplement & +16 documents & +32 documents & +64 documents & Best size \\
\midrule
ARC/OBQA-train  & +1.27 & \textbf{+1.60} & +1.19 & 32 \\
MMLU-aux-train  & +0.27 & +0.39 & \textbf{+0.64} & 64 \\
HellaSwag-train & \textbf{+0.77} & +0.30 & +0.49 & 16 \\
\bottomrule
\end{tabular}
\end{table}

\subsection{Cross-family transfer matrix}

The seven-task ``Other'' group includes ARC-Easy and OpenBookQA.
It is distinct from the five-task strict non-targeted group in
\cref{tab:targeting}.

\begin{table}[H]
\centering
\small
\caption{SR-V2 W2G128 task-wise gains (pp, five-model mean, seed 42).
Selected doses use the runs and Pile-128 controls in \cref{tab:fragcal_dose_response}.
AVG averages ten tasks. Other (7) excludes MMLU, ARC-C, and HellaSwag but includes ARC-E/OBQA.}
\label{tab:cross_family}
\begin{minipage}{\linewidth}
\centering
\begin{tabular}{@{}lccccc@{}}
\toprule
Supplement & MMLU & ARC-C & HellaSwag & Other (7) & $\Delta$AVG \\
\midrule
MMLU$\times$32      & \textbf{+5.65}$^\dagger$ & $-$0.61 & +0.76 & $-$0.27 & +0.39 \\
ARC/OBQA$\times$32  & \textbf{+6.87} & \textbf{+2.81}$^\dagger$ & +0.54 & +0.82 & +1.60 \\
HellaSwag$\times$16 & +0.30 & +1.07 & \textbf{+3.30}$^\dagger$ & +0.43 & +0.77 \\
\bottomrule
\end{tabular}
\\[2pt]
{\footnotesize $^\dagger$In-family (supplement targets this task).}
\end{minipage}
\end{table}

The seed-42 comparison favors ARC/OBQA over MMLU supplements on
MMLU in four of five models. Both pools contain multiple-choice QA, so the
comparison does not isolate format or activation coverage. The four-seed
replication is reported in \cref{app:transfer}.

\subsection{Final recipe}

The supplement sizes are the across-model medians of the five
model-specific dose recommendations. They allocate 32 MMLU, 32 ARC/OBQA,
and 16 HellaSwag documents. These recommendations consider target-task
saturation and non-target-task degradation. They do not maximize the
pooled ten-task mean, so the final sizes need not equal
the Best size entries in \cref{tab:fragcal_dose_response}.

\begin{table}[H]
\centering
\small
\caption{Final FragCal-DC allocation of 208 raw documents before
tokenization and packing into fixed-length sequences.}
\label{tab:fragcal_recipe}
\begin{tabularx}{\linewidth}{@{}llcX@{}}
\toprule
Component & Source & Documents & Purpose \\
\midrule
Pile anchor     & pile-10k         & 128 & General coverage \\
MMLU supplement & auxiliary\_train & 32  & Reasoning/knowledge coverage \\
ARC/OBQA supplement & ARC-C/E + OBQA train & 32 & Task-aligned QA coverage \\
HS supplement   & HellaSwag train  & 16  & Commonsense coverage \\
\midrule
\textbf{Total}  &                  & \textbf{208} & Raw documents; effective sequence counts depend on tokenizer and loader \\
\bottomrule
\end{tabularx}
\end{table}

The four seed variants are FragCal-DC-s42, s123, s456, and s789.

\section{FragCal-DC reproduction script}
\label{app:fragcal_script}

The supplementary material provides \texttt{prep\_fragcal.py} and protocol
documentation under Apache-2.0, without datasets or model weights.
This appendix explains how to generate and identify the calibration
artifacts. Construction choices are detailed in \cref{app:fragcal}.

\subsection{Artifact modes and checksums}
\label{app:fragcal_script_checksums}

The script supports three mutually exclusive answer-rendering modes
via command-line flags, each producing a distinct family of
calibration artifacts:

\begin{itemize}
  \item \textbf{FragCal-DC} (default): each MMLU/ARC/OBQA row ends
        with \texttt{Answer: X)~\textit{text}} and HellaSwag rows
        append the correct ending to the context. This is the
        primary calibration set used throughout the main paper.
  \item \textbf{FragCal-DC (wrong answers)} (\texttt{-{}-wrong-answers}):
        the ablation in \cref{sec:wrong_answer}, in which each MMLU/ARC/OBQA
        answer is replaced by a uniformly-sampled incorrect
        alternative and each HellaSwag row appends a
        uniformly-sampled incorrect ending. The Pile slice is
        unchanged.
  \item \textbf{FragCal-DC (no answer designated)} (\texttt{-{}-no-answers}):
        the no-designation ablation in \cref{sec:wrong_answer}, in which the
        \texttt{Answer:} line is removed from MMLU/ARC/OBQA rows
        and HellaSwag rows are rendered as an all-endings
        ``Possible continuations'' list with no selection made.
        The Pile slice is unchanged.
\end{itemize}

Run the script once per seed, for example
\texttt{python prep\_fragcal.py -{}-seed 42}, and repeat for seeds 123,
456, and 789. Add the relevant ablation flag for the answer controls.
The public HuggingFace inputs are pinned in \cref{app:fragcal_script_deps}.
The SHA-256 inventory below covers four default files, the seed-42
wrong-designation file, and four no-designation files.

{\scriptsize
\begin{verbatim}
# FragCal-DC (default)
bea2cbdc5c8cc3cdea82882bbe7b4f8e636c2ade1f0751e8896e10fc5994563e  fragcal_s42.json
870cf93ec25bffd2742abee0b0200f88d75c8d756346b54465a15fa648a82cef  fragcal_s123.json
06d098099a53de3c93d907ae250afb7f78f50a7d9c194fdeddffd8cb48fb0fc5  fragcal_s456.json
160ab5db52981cd73430811906b20edb09b2d8c170b7284d3d7937acb5cae1d5  fragcal_s789.json

# FragCal-DC (wrong answers)  --  --wrong-answers
60d7262c288cfaa589125ea4b35fd13e98fa4e2aa76e31862ac5f0030945eafe  fragcal_s42_wrong.json

# FragCal-DC (no answer designated)  --  --no-answers
718eb37515f4b62d513b272cd4e3f6c9b983dce6456ab940fc1837d607ccc99a  fragcal_s42_noans.json
17b551ff80545152bcd8f9aafbcad5dc1bdf1089fdab3ddf1d2b0082e71a76a6  fragcal_s123_noans.json
e2b11882ffa9710b4c9388865ff06fb64a88a694070f7d6f3627b402b275913a  fragcal_s456_noans.json
1bde0be5b9c2201fe4bada32eda4e840fc93df2e7932cdaa5025ebaa01eaba53  fragcal_s789_noans.json
\end{verbatim}
}

Each file contains 208 entries of the form
\texttt{\{"text": "..."\}}. Pile entries are kept whole and contain
at least 8{,}000 characters. QA packing uses a 12{,}000-character running
target. Counting a trailing separator can leave the joined text two
characters shorter (minimum 11{,}998 in the listed files).
Character counts do not guarantee a 2048-token sequence.
The recipe, seeds, and
upstream dataset revisions (\cref{app:fragcal_script_deps}) define the
construction inputs. The hashes identify the calibration artifacts used
in the study.

\subsection{Dependencies and pinned revisions}
\label{app:fragcal_script_deps}

The script requires \texttt{datasets} version 2.14 or later and
pulls the following public HuggingFace datasets at runtime. The
\texttt{revision} argument to \texttt{load\_dataset} pins each
source to a specific commit SHA on the HuggingFace hub, fixing the dataset
contents independently of later upstream changes:

{\scriptsize
\begin{verbatim}
NeelNanda/pile-10k             127bfedcd5047750df5ccf3a12979a47bfa0bafa
cais/mmlu                      c30699e8356da336a370243923dbaf21066bb9fe
allenai/ai2_arc                210d026faf9955653af8916fad021475a3f00453
allenai/openbookqa             388097ea7776314e93a529163e0fea805b8a6454
Rowan/hellaswag                218ec52e09a7e7462a5400043bb9a69a41d06b76
\end{verbatim}
}

The calibration files are derived from these sources. Their use and
redistribution remain subject to the respective source terms.
Source pins do not freeze the Python and \texttt{datasets} execution
environment. Generated files should be checked against the listed hashes.

\subsection{Supplementary implementation}
\label{app:fragcal_script_src}

The supplementary file \texttt{prep\_fragcal.py} contains the complete source.
The mutually exclusive flags \texttt{-{}-wrong-answers} and
\texttt{-{}-no-answers} select the two answer ablations documented
above. Per-row formatters
(\texttt{format\_mmlu}, \texttt{format\_arc}, \texttt{format\_obqa},
\texttt{format\_hellaswag}) construct each row's rendered text via a
shared \texttt{mode} argument. The function \texttt{\_concat\_to\_long\_docs}
packs short Q\&A items using the 12{,}000-character running target
before subsampling to the target count.

\subsection{Asset sources and terms}
\label{app:asset_terms}

The supplementary \texttt{asset\_terms.csv} identifies the model, dataset,
and principal software sources and their declared terms. Qwen3,
Qwen2.5-7B-Instruct, Mistral-7B-v0.1, and Mixtral-8x7B-Instruct-v0.1
declare Apache-2.0; Phi-3 declares MIT. Llama-2 and Llama-3.1 use their
respective Community License Agreements, and Gemma-2 uses the Gemma Terms
of Use. These model-specific terms must not be described as a uniform
unrestricted open-source license.

The pinned calibration cards declare MIT for MMLU and HellaSwag,
CC-BY-SA-4.0 for ARC, and \texttt{bigscience-bloom-rail-1.0} for the
Pile-10k distribution. OpenBookQA's pinned card marks its license as
unknown, while its upstream repository provides Apache-2.0 terms.
PIQA's author distribution declares AFL-3.0, GPQA's
dataset card declares CC BY 4.0, and OmniQuant's upstream implementation
declares MIT.

The authors confirm compliance with the applicable asset licenses and
terms for the reported research; no license conflicts were identified.
The supplementary construction code and documentation are released under
Apache-2.0. This license covers the authors' contributions; upstream
assets retain their respective terms. No datasets or model weights are
included in the supplementary material.

\section{Hardware and implementation details}
\label{app:hardware}

\subsection{Platform details}

All experiments used Intel Data Center GPU Max 1550 (Ponte Vecchio, PVC), with 6 GPUs per node
and 128~GB HBM2e per GPU (64~GB per tile, dual-stack).
Software versions were PyTorch 2.10.0+xpu, transformers 5.5.0, and lm-eval 0.4.11.

The platform used copy-engine (BCS) driver workarounds, single-tile execution through
\texttt{ZE\_AFFINITY\_MASK=0}, and the native memory allocator.
Method-specific changes were as follows.
\begin{itemize}[itemsep=1pt]
  \item \textbf{GPTQModel.} Cholesky factorization offloaded to the CPU for XPU NaN failures,
        disabled \texttt{torch.compile}, and backend selection.
  \item \textbf{AWQ.} CUDA$\to$XPU device abstraction and Mistral/Qwen3 support.
  \item \textbf{OmniQuant.} A standalone LWC-only harness for XPU.
  \item \textbf{AQLM.} Codebook training ported to PVC tiles.
\end{itemize}

\subsection{Implementation and evaluation settings}
\label{app:implementation}

Recorded calibration sequence lengths are 512 tokens for AWQ/OmniQuant and
2048 for GPTQ/SR-V2/RTN-opt.

The dense GPTQ checkpoints identify GPTQModel 6.1.0-dev and use symmetric
integer quantization, no activation ordering, sequential quantization,
damping 0.05 with 0.01 increments, non-static groups, and
activation-group-aware processing. AWQ uses zero points and a fake-quantized
checkpoint backend. Phi-3 uses BF16 and the other five dense models FP16.
OmniQuant uses the LWC-only fake-quantized backend with 20 epochs and
clipping learning rate 0.01.

SR-V2 uses AutoRound 0.12.0 with the algorithm extension enabled and
200 iterations. For Pile-200 and FragCal, its saved batch-size field is
present in 72 of 144 dense configurations, with values 1 or 4.
These records do not define a universal
batch size. RTN-opt's W3/W4 configurations identify the same package version
and zero iterations. AQLM's $1{\times}16g8$ representation uses one 16-bit-index codebook, input/output
groups 8/1, \texttt{scale\_nbits=0}, sequence length 2048, up to 100 optimization
epochs, beam size 1, relative-MSE stopping tolerance 0.01, and up to five
fine-tuning epochs. Calibration-count differences are specified in
\cref{app:calibration_accounting}.

The ten-task evaluation wrapper selects length-normalized accuracy for
ARC-Challenge, HellaSwag, and OpenBookQA, and accuracy for the other tasks.
TruthfulQA uses MC1. Evaluation-harness random seeds are distinct from
calibration seeds. The recorded settings are Python 0 and NumPy, Torch,
and few-shot sampling 1234. The wrapper uses the selected harness task
definitions without a global few-shot override. Llama SR-V2 evaluation
explicitly enables a beginning-of-sequence (BOS) token. Package-version labels do not uniquely
identify local XPU patches. The supplementary release provides the calibration
generator and protocol; the complete method-specific XPU implementations and
execution environments are not included.

\subsection{End-to-end reproduction workflow}
\label{app:reproduction}

The main comparisons follow four stages. The settings above and
\cref{sec:setup,app:calibration_accounting} specify the method-dependent
configuration and sample accounting.

\begin{enumerate}[leftmargin=*]
  \item \textbf{Prepare calibration samples.}
        For each calibration seed in $\{42,123,456,789\}$, prepare the specified
        Pile control and generate FragCal-DC with the supplementary
        \texttt{prep\_fragcal.py} script (\cref{app:fragcal_script}).
        Use Pile-200 for the common-reference width comparisons and retain
        the explicitly identified Pile-128 controls in the other analyses.
        Preserve the 208-document FragCal recipe, source splits, and packing
        rules. Apply each method's sequence length and sample cap as
        documented; effective loader counts are reported in
        \cref{app:calibration_accounting}.
  \item \textbf{Run paired quantization.}
        For each model, method, bit width, and seed, start both calibration
        conditions from the same unquantized model. Change the calibration
        samples while holding the quantizer configuration fixed. Use group
        size 128 for scalar quantizers and the stated W2-equivalent AQLM
        representation. Preserve the LWC-only OmniQuant setting, SR-V2's
        algorithm extension, and the remaining options above. Save each
        resulting checkpoint together with its calibration condition and
        quantization configuration.
  \item \textbf{Evaluate checkpoints.}
        Load each checkpoint with its method-compatible backend and run
        lm-eval 0.4.11 on the ten tasks in \cref{sec:setup}.
        Within each calibration pair, keep the tokenizer, model-loading
        precision, BOS handling, task definitions, and evaluation seeds fixed.
        Use the harness task definitions' few-shot settings and the
        task-specific accuracy metrics listed above. Evaluate the unquantized
        reference with the corresponding task and scoring settings.
  \item \textbf{Aggregate paired results.}
        Compute the unweighted ten-task accuracy for each run, then the
        FragCal minus control difference within each seed. Average the four
        paired differences within each model before summarizing across
        models. Apply the stated functional-model selection for OmniQuant
        and report its all-model sensitivity separately. Use the model-level
        intervals and sign-flip tests in \cref{sec:setup}; retain the distinct
        references and gap-closure definitions for the matched-width and
        original descriptive comparisons (\cref{sec:bitwidth,app:matched}).
\end{enumerate}

\subsection{Compute budget}

Approximate single-run wall times use a single PVC tile and a 7B model at
W2G128, with W2-equivalent storage for AQLM.
They are RTN $<$1~min, RTN-opt $\sim$5~min, AWQ $\sim$20~min,
GPTQ $\sim$45~min, OmniQuant $\sim$35~min, SignRound~V2 $\sim$35~min,
and AQLM $\sim$6--24~h.
Applying FragCal requires one re-quantization. The approximate budget
quoted for SignRound~V2 on a 7B model is $\sim$1~hour.

\paragraph{Total compute estimate.}
The authors estimate the overall research compute budget at approximately
3,000 PVC tile-hours, including preliminary experiments.
One tile-hour corresponds to one PVC tile used for one hour; each physical
PVC GPU has two tiles. This is an approximate resource-use budget.
The representative per-run costs above describe variation across methods.
The evaluation subtotal below covers part of this workload and is already
included in the estimate.

\paragraph{Recorded evaluation subtotal.}
For the dense W4 comparison, matched timing records cover 142 of 144
evaluations, comprising 48 AWQ, 46 GPTQ, and 48 OmniQuant runs. Summed evaluation-process
wall times are 21.93, 108.82, and 24.05 hours, respectively
(154.80 hours total). Per-run ranges are 16.0--34.2, 46.5--189.7, and
18.4--42.5 minutes. These include loading and I/O within the evaluator,
not just device kernels. They are not GPU-utilization hours or elapsed
project time. The subtotal excludes two GPTQ runs without unambiguous
matching timing lines, all quantization time, other widths and methods,
scale/architecture extensions, and failed or exploratory runs. It is
not a study-wide compute estimate.

\subsection{Gemma-2-9B architecture probe}
Gemma-2-9B~\citep{riviere2024gemma2} is a separate architecture probe and is excluded from every primary
aggregate. The following implementation outcomes limit its interpretation.

\textbf{AQLM} was not extended to Gemma-2-9B. Its beam-search calibration
($k$-means codebook initialization plus 100 beam epochs $\times$ 7 linears
$\times$ 42 layers) projects to ${\sim}$270 GPU-hours per
(model, calibration) pair, exceeding the 24-hour PBS job limit.
Reducing beam epochs would change the protocol used for the other five architectures.

\textbf{AWQ} (AutoAWQ $0.2.8$) produced degenerate Gemma evaluations at
all three widths and both calibration conditions over four seeds. The repeated
TruthfulQA MC1 score of 1.0 together with zero LAMBADA accuracy makes these
outputs unsuitable for assessing quantized-model accuracy. The responsible
implementation or architecture interaction has not been isolated.

\textbf{OmniQuant} has four-seed W2/W3/W4 coverage. Its standard W2
five-task diagnostic mean declines from 0.6332 with Pile to 0.5770 with
FragCal, a $-5.62$ pp response. This negative result limits generalization
beyond the primary panel.

\textbf{SignRound~V2, GPTQ, and RTN-opt} also remain outside the primary
aggregates on Gemma, regardless of their individual compatibility.

The seven LWC hyperparameter probes use FragCal seed 42.
At sequence length 512, increasing training from 5 to 40 epochs raises the
five-task mean from 0.5079 to 0.5712. At 20 epochs, increasing context from
1024 to 2048 raises it from 0.5537 to 0.5698. The probes support sensitivity to optimization settings,
but are not paired against the standard four-seed references and do not
intervene on soft-capping, normalization layout, or sliding-window attention.

\section{Paired responses and seed variability}
\label{app:seed_variance}

Seed-level results show variation across repeated calibration runs.
The first table reports the complete five-model Pile-200 task-level panel for
SR-V2 and RTN-opt at W2G128. It contains forty paired records, with seeds
42, 123, 456, and 789 for each method-model combination. The four deltas
are averaged within each method-model pair before calculating the main-text interval.
Seed variability and cross-model uncertainty are different quantities.
Small seed variability does not establish a class effect.

{\small
\begin{longtable}{@{}llcrrr@{}}
\caption{W2G128 ten-task mean accuracies (AVG, fractions) by calibration
seed. Pile denotes Pile-200. $\Delta$ is FragCal minus Pile (pp).}\\
\toprule
Method & Model & Seed & Pile & FragCal & $\Delta$ pp\\
\midrule\endfirsthead
\multicolumn{6}{@{}l}{Table \thetable\ (continued)}\\
\toprule
Method & Model & Seed & Pile & FragCal & $\Delta$ pp\\
\midrule\endhead
\bottomrule\endfoot
\inputrows{generated/seed_rows.tex}
\end{longtable}
}

The cross-method table in \cref{tab:cross_models} reports four-seed aggregates
for the other methods.

\subsection{Additional seed panels}

The additional seed panels below are separate measurements. Their GPTQ
baseline scores differ from the paired-panel means and are not substituted
for those controls. Means, seed standard deviations, and CVs are descriptive
summaries, not confidence intervals for a calibration effect. The rounded
seed scores do not specify a common SD divisor across these panels.

\begin{table}[H]
\centering\small
\caption{Additional GPTQ W3G128 generic-Pile seed panel. Entries are
ten-task mean accuracies (fractions). SD is seed standard deviation
and CV is $100\,\mathrm{SD}/\mathrm{Mean}$ (\%).}
\label{tab:seed_gptq}
\begin{tabular}{@{}lrrrrrrr@{}}
\toprule
Model & s42 & s123 & s456 & s789 & Mean & SD & CV\\
\midrule
\inputrows{generated/historical_seed_gptq_rows.tex}
\bottomrule
\end{tabular}
\end{table}

\begin{table}[H]
\centering\small
\caption{Additional SR-V2 W2G128 FragCal ten-task mean accuracies
(fractions), not paired differences. SD is seed standard deviation.
CV is $100\,\mathrm{SD}/\mathrm{Mean}$ (\%).}
\label{tab:seed_srv2}
\begin{tabular}{@{}lrrrrrrr@{}}
\toprule
Model & s42 & s123 & s456 & s789 & Mean & SD & CV\\
\midrule
\inputrows{generated/historical_seed_srv2_rows.tex}
\bottomrule
\end{tabular}
\end{table}

The below-threshold OmniQuant Mistral configuration contains an anomalous
TruthfulQA score. Its four-seed mean in \cref{tab:cross_models} is reported
as observed, without imputing a replacement score. The configuration is
excluded from OmniQuant's five-model response estimate.
The descriptive panel below separately reports the three-seed Pile summary.

\begin{table}[H]
\centering\small
\setlength{\tabcolsep}{3pt}
\caption{OmniQuant W2G128 ten-task accuracy across seeds.
SD is standard deviation, CV is in percent, and range is max minus min.
Mistral's Pile row ($\dagger$) excludes anomalous seed 456, using three seeds
instead of four in \cref{tab:cross_models}.}
\label{tab:seed_omniquant_w2}
\begin{tabular}{@{}lrrrrrrrr@{}}
\toprule
& \multicolumn{4}{c}{Pile} & \multicolumn{4}{c}{FragCal}\\
\cmidrule(lr){2-5}\cmidrule(lr){6-9}
Model & Mean & SD & CV & Range & Mean & SD & CV & Range\\
\midrule
\inputrows{generated/historical_seed_oq_rows.tex}
\bottomrule
\end{tabular}
\end{table}

\begin{table}[H]
\centering\small
\caption{Typical seed standard deviations by method and bit width
(ten-task mean accuracy, fractions). These are variability summaries,
not uncertainty intervals for a calibration effect.}
\label{tab:seed_summary}
\begin{tabular}{@{}lrrr@{}}
\toprule
Method & W2G128 & W3G128 & W4G128\\
\midrule
\spclass~GPTQ & --- & 0.0024 & ---\\
\itclass~OmniQuant & 0.0031 & 0.0023 & 0.0011\\
\itclass~SR-V2 & 0.0030 & --- & ---\\
\bottomrule
\end{tabular}
\end{table}

\section{Calibration objectives and reconstruction error}
\label{app:calibration_loss}

Method-specific objectives describe optimization on each calibration set.
The fixed-reference panel holds evaluation inputs constant to compare
reconstruction changes with task accuracy (\cref{sec:loss_paradox}).

For the recorded objectives, relative change is
$\Delta(\%)=100(L_{\rm FragCal}/L_{\rm Pile}-1)$, so negative
values indicate a lower objective. Displayed losses are rounded.
Their ratios need not reproduce the reported deltas exactly. The objective
summaries and detailed panels have different coverage and are not pooled.

\begin{table}[H]
\centering\small
\caption{Method-specific calibration-objective changes at the operating widths.
Inputs differ across calibrations, so these are not common-reference losses.
RTN-opt reports calibration-perplexity change. See \cref{tab:heldout_permodel} for fixed-reference results.}
\label{tab:calibration_proxy_means}
\begin{tabular}{@{}lcr@{}}
\toprule
Method & Width & Recorded objective change (\%)\\
\midrule
\spclass~RTN-opt & W2G128 & $-0.05$\\
\spclass~AWQ & W3G128 & $+7.80$\\
\spclass~GPTQ & W3G128 & $-75.13$\\
\midrule
\itclass~OmniQuant & W2G128 & $-9.31$\\
\itclass~SR-V2 & W2G128 & $-3.74$\\
\itclass~AQLM & W2-equivalent & $+5.63$\\
\bottomrule
\end{tabular}
\end{table}

\subsection{GPTQ reconstruction error (Hessian-weighted)}

\begin{table}[H]
\centering
\small
\caption{GPTQ Hessian-weighted calibration loss, group size 128.
$\Delta$ is relative change from Pile-200 to FragCal (\%, negative is lower).
Losses use different calibration inputs.}
\label{tab:calibloss_gptq_all}
\begin{tabular}{@{}lcrrr@{}}
\toprule
Model & Bit & Pile-200 Loss & FragCal Loss & $\Delta$ (\%) \\
\midrule
Qwen3-1.7B  & W2G128 & 8.892 & 2.639 & $-$70.3\% \\
Qwen3-1.7B  & W3G128 & 2.007 & 0.528 & $-$73.7\% \\
Qwen3-1.7B  & W4G128 & 0.463 & 0.122 & $-$73.7\% \\
Llama-2-7B   & W2G128 & 0.249 & 0.064 & $-$74.2\% \\
Mistral-7B   & W2G128 & 0.074 & 0.000 & $-$99.4\% \\
Q2.5-7B-Inst & W2G128 & 0.640 & 0.173 & $-$73.0\% \\
Qwen3-8B     & W2G128 & 0.758 & 0.267 & $-$64.8\% \\
\bottomrule
\end{tabular}
\end{table}

\subsection{GPTQ W3 reconstruction error (Hessian-weighted)}

\begin{table}[H]
\centering
\small
\caption{GPTQ W3G128 Hessian-weighted calibration loss by model.
$\Delta$ is relative change from Pile-200 to FragCal (\%, negative is lower).
The Mistral entry covers only non-fallback modules.}
\label{tab:calibloss_gptq_w3}
\begin{minipage}{\linewidth}
\centering
\begin{tabular}{@{}lcrrr@{}}
\toprule
Model & Bit & Pile-200 Loss & FragCal Loss & $\Delta$ (\%) \\
\midrule
Qwen3-1.7B             & W3G128 & 2.007 & 0.528 & $-$73.7\% \\
Llama-2-7B             & W3G128 & 0.043 & 0.012 & $-$72.5\% \\
Mistral-7B             & W3G128 & 0.008 & 0.000 & $-$96.9\%$^\dagger$ \\
Q2.5-7B-Inst           & W3G128 & 0.114 & 0.031 & $-$72.7\% \\
Qwen3-8B               & W3G128 & 0.162 & 0.058 & $-$64.4\% \\
Phi-3                  & W3G128 & 0.103 & 0.030 & $-$70.5\% \\
\bottomrule
\end{tabular}
\\[2pt]
{\footnotesize $^\dagger$Mistral-7B at W3 hits Hessian-collapse RTN fallback on 11--91\% of layer-modules per seed (Pile) and 71--91\% (FragCal); the residual GPTQ loss covers only non-fallback rows. Excluding Mistral, the panel range is $-$64\% to $-$74\% (dense, five models).}
\end{minipage}
\end{table}

\subsection{Fixed-reference reconstruction comparison}

The comparison evaluates reconstruction on a fixed
128-sample Pile reference at seed 999. Negative MSE change is better. Positive
accuracy change is better. A directional mismatch therefore occurs when the
two nonzero deltas have the same algebraic sign. All controls use Pile-200.
MSE changes average the available paired quantization seeds within each model.
These are 42/123/456/789 for GPTQ and AWQ, seed 123 for SR-V2, and seed 42 for
OmniQuant and AQLM, except their Phi-3 cells use all four seeds.
Accuracy changes use the corresponding four-seed model means.
This is a descriptive model-level comparison, not a run-level correlation.
The reference uses held-out Pile text. Behavior on an independent
out-of-distribution corpus is untested.

{\small
\begin{longtable}{@{}llrr@{}}
\caption{Fixed-reference reconstruction change (\%) and downstream accuracy
change (pp), both FragCal minus Pile. GPTQ/AWQ use W3G128,
SR-V2/OmniQuant use W2G128, and AQLM is
W2-equivalent. OmniQuant's below-threshold Mistral cell is omitted.
Accuracy changes are rounded from the paired-score records.}
\label{tab:heldout_permodel}\\
\toprule
Method & Model & $\Delta$ MSE (\%) & $\Delta$ accuracy (pp)\\
\midrule\endfirsthead
\multicolumn{4}{@{}l}{Table \thetable\ (continued)}\\
\toprule
Method & Model & $\Delta$ MSE (\%) & $\Delta$ accuracy (pp)\\
\midrule\endhead
\bottomrule\endfoot
\spclass~AWQ & Qwen3-1.7B & $+4.36$ & $-0.35$\\
 & Llama-2-7B & $+10.12$ & $+0.07$\\
 & Mistral-7B & $+5.95$ & $+0.25$\\
 & Qwen2.5-7B & $+22.77$ & $-0.49$\\
 & Qwen3-8B & $+2.24$ & $+0.15$\\
 & Phi-3 & $+1.33$ & $+0.14$\\
\midrule
\spclass~GPTQ & Qwen3-1.7B & $+12.12$ & $-0.84$\\
 & Llama-2-7B & $+9.91$ & $-0.25$\\
 & Mistral-7B & $+5.02$ & $-0.98$\\
 & Qwen2.5-7B & $+1.53$ & $+0.63$\\
 & Qwen3-8B & $-6.34$ & $+1.62$\\
 & Phi-3 & $+1.18$ & $+1.21$\\
\midrule
\itclass~OmniQuant & Qwen3-1.7B & $+1.47$ & $+1.89$\\
 & Llama-2-7B & $-6.04$ & $+1.27$\\
 & Qwen2.5-7B & $+3.21$ & $+1.53$\\
 & Qwen3-8B & $-1.58$ & $+2.32$\\
 & Phi-3 & $+1.82$ & $+1.87$\\
\midrule
\itclass~SR-V2 & Qwen3-1.7B & $-2.81$ & $+2.31$\\
 & Llama-2-7B & $-11.48$ & $+1.31$\\
 & Mistral-7B & $+0.33$ & $+2.36$\\
 & Qwen2.5-7B & $+8.82$ & $+0.54$\\
 & Qwen3-8B & $-1.31$ & $+1.50$\\
 & Phi-3 & $-24.25$ & $+2.97$\\
\midrule
\itclass~AQLM & Qwen3-1.7B & $+6.18$ & $+1.76$\\*
 & Llama-2-7B & $+1.42$ & $+0.58$\\*
 & Mistral-7B & $+3.43$ & $+1.21$\\*
 & Qwen2.5-7B & $+20.58$ & $-0.71$\\*
 & Qwen3-8B & $+0.47$ & $+0.59$\\*
 & Phi-3 & $+2.46$ & $+0.98$\\
\end{longtable}
}

\subsection{SignRound~V2 convergence loss}

\begin{table}[H]
\centering
\small
\caption{SR-V2 W2G128 final calibration loss.
$\Delta$ is relative change from Pile-200 to FragCal (\%, negative is lower).}
\label{tab:calibloss_srv2}
\begin{tabular}{@{}lcrrr@{}}
\toprule
Model & Bit & Pile-200 Final & FragCal Final & $\Delta$ (\%) \\
\midrule
Qwen3-1.7B  & W2G128 & 939.08 & 872.90 & $-$7.0\% \\
Llama-2-7B   & W2G128 & 2.16   & 2.52   & +16.7\% \\
Mistral-7B   & W2G128 & 0.11   & 0.11   & $\phantom{+}$0.0\% \\
Q2.5-7B-Inst & W2G128 & 17.92  & 17.39  & $-$3.0\% \\
Qwen3-8B     & W2G128 & 76.80  & 73.07  & $-$4.9\% \\
\bottomrule
\end{tabular}
\end{table}

\subsection{OmniQuant LWC loss (20-epoch final)}

\begin{table}[H]
\centering
\small
\caption{OmniQuant W2G128 LWC calibration loss after 20 epochs.
$\Delta$ is relative change from Pile-200 to FragCal (\%, negative is lower).}
\label{tab:calibloss_omniquant}
\begin{minipage}{\linewidth}
\centering
\begin{tabular}{@{}lcrrr@{}}
\toprule
Model & Bit & Pile-200 Loss & FragCal Loss & $\Delta$ (\%) \\
\midrule
Qwen3-1.7B  & W2G128 & 741.5 & 629.6 & $-$15.1\% \\
Llama-2-7B   & W2G128 & 2.54  & 2.34  & $-$7.7\% \\
Mistral-7B   & W2G128 & 0.187 & 0.215 & +15.0\%$^*$ \\
Q2.5-7B-Inst & W2G128 & 19.6  & 17.3  & $-$11.7\% \\
Qwen3-8B     & W2G128 & 68.5  & 60.6  & $-$11.5\% \\
\bottomrule
\end{tabular}
\\[2pt]
{\footnotesize $^*$Mistral includes the anomalous Pile seed-456 record.
Its below-threshold W2 baseline excludes it from the five-model response
estimate. The recorded losses are retained without imputation.}
\end{minipage}
\end{table}

\section{Optimizer changes and repeated quantization}
\label{app:dm_signsgd}

Two interventions test whether further reconstruction-loss reduction
improves task accuracy (\cref{sec:loss_paradox}).

\subsection{DM-SignSGD vs.\ SignSGD}

We replace SignRound~V2's default SignSGD with a dual-momentum variant
(DM-SignSGD), retaining the quantization configuration.

\begin{table}[H]
\centering
\small
\caption{DM-SignSGD changes relative to SignSGD within each model and configuration.
AVG is ten-task accuracy (fraction), $\Delta$AVG is in pp, and $\Delta$Loss is relative percent.
GC is the percentage of the gap between SignSGD and full-precision accuracy recovered by DM-SignSGD.}
\begin{tabular}{@{}llcrrrrr@{}}
\toprule
Config & Model & Optimizer & Loss & $\Delta$Loss & AVG & $\Delta$AVG & GC\% \\
\midrule
W2G16 & Q3-1.7B & SignSGD & 465.8 & --- & .5025 & --- & --- \\
      & & DM-SignSGD & 360.3 & $-$22.6\% & .5082 & +0.57 & +5.9\% \\
\midrule
W2G16 & Q2.5-7B & SignSGD & 9.72 & --- & .6920 & --- & --- \\
      & & DM-SignSGD & 6.83 & $-$29.7\% & .6901 & $-$0.19 & $-$3.8\% \\
\midrule
W2G128 & Q3-1.7B & SignSGD & 790.1 & --- & .4295 & --- & --- \\
       & & DM-SignSGD & 639.8 & $-$19.0\% & .4297 & +0.02 & +0.2\% \\
\midrule
W2G128 & Q2.5-7B & SignSGD & 14.2 & --- & .6031 & --- & --- \\
       & & DM-SignSGD & 13.3 & $-$6.3\% & .5962 & $-$0.69 & $-$8.0\% \\
\bottomrule
\end{tabular}
\end{table}

Calibration loss falls by 6--30\% in all four configurations, whereas
accuracy changes range from $-0.69$ to $+0.57$ pp. Qwen2.5-7B at W2G128
loses $0.69$ pp despite a 6.3\% loss reduction. This mismatch supports direct
accuracy checks when changing the optimizer. These measurements do not
identify an overfitting mechanism or test preservation of gradient directions.

\subsection{Multi-pass refinement}

The second quantization pass starts from the first pass's quantized
checkpoint, changing the reconstruction problem. The panel
contains five models (Llama-2-7B, Mistral-7B, Qwen2.5-7B-Instruct,
Qwen3-1.7B, and Qwen3-8B), two calibrations, and two bit widths,
giving twenty comparisons. Phi-3 is not included.

\begin{table}[H]
\centering
\small
\caption{SR-V2 second-pass refinement over five models, two calibrations, and two widths ($n=20$ cells).
Loss is normalized to pass 1.
$\Delta$AVG is pass-2 minus pass-1 ten-task accuracy (pp).}
\begin{tabular}{@{}lcc@{}}
\toprule
Quantity & Pass~1 & Pass~2 \\
\midrule
Mean reconstruction loss (relative) & $1.0\times$ & $\sim 10^{-4}\times$ \\
Downstream $\Delta$AVG (pp), median over 20 cells & ---   & $-0.26$ \\
Cells with $\Delta$AVG $< 0$ & ---   & 15/20 \\
Cells with $\Delta$AVG $\geq +0.5$pp & ---   & 1/20 \\
\bottomrule
\end{tabular}
\end{table}

Pass~2 achieves $10{,}000$--$12{,}000\times$ lower recorded reconstruction
loss, while 15 of 20 downstream deltas are negative (median $-0.26$ pp).
Only one cell improves by at least $0.5$ pp. These results show that
the lower second-pass objective is not a reliable selection criterion for
these configurations.

\section{Iteration-count comparisons}
\label{app:iteration_ablation}

The sweep examines how SignRound~V2 accuracy and calibration response vary
with iteration budget (\cref{sec:wrong_answer}).
It covers iteration count $i \in \{10, 50, 100, 200\}$ across all 6 models,
both calibration conditions, and 4 seeds. The models are Llama-2-7B,
Mistral-7B, Qwen2.5-7B-Inst, Qwen3-1.7B, Qwen3-8B, and Phi-3-mini-4k-instruct.

The available five-model 10/50/100-step checkpoints are W2G128.
Their displayed spreads use sample SD (variance divisor $4-1=3$).
Their historical metric selection additionally prefers normalized accuracy
for ARC-Easy, PIQA, and WinoGrande when provided, with ordinary accuracy
as fallback. This differs from the primary panel. The 200-step references
and Phi-3 results have not been fully reconciled to that selection.
The tables are therefore historical summaries, not a protocol-harmonized
comparison across all iteration counts.

\subsection{Generic-calibration accuracy}

\begin{table}[H]
\centering
\small
\caption{SR-V2 Pile-128 ten-task mean accuracies (fractions) by iteration
count, reported as four-seed mean $\pm$ seed SD.
$\Delta$ is final minus initial accuracy, also in fractions.}
\label{tab:iter_pile_accuracy}
\begin{tabular}{@{}lccccc@{}}
\toprule
Model & $i{=}10$ & $i{=}50$ & $i{=}100$ & $i{=}200$ & $\Delta(10{\to}200)$ \\
\midrule
Q3-1.7B  & .290 $\pm$ .002 & .408 $\pm$ .003 & .419 $\pm$ .007 & .430 $\pm$ .004 & +.139 \\
Q2.5-7B  & .322 $\pm$ .033 & .568 $\pm$ .004 & .589 $\pm$ .002 & .603 $\pm$ .002 & +.282 \\
Ll2-7B   & .374 $\pm$ .033 & .502 $\pm$ .003 & .513 $\pm$ .002 & .526 $\pm$ .002 & +.153 \\
Mi-7B    & .428 $\pm$ .016 & .541 $\pm$ .007 & .553 $\pm$ .005 & .560 $\pm$ .002 & +.132 \\
Q3-8B    & .304 $\pm$ .021 & .547 $\pm$ .004 & .559 $\pm$ .002 & .570 $\pm$ .002 & +.266 \\
Phi-3    & .340 $\pm$ .028 & .494 $\pm$ .005 & .519 $\pm$ .001 & .521 $\pm$ .004 & +.181 \\
\bottomrule
\end{tabular}
\end{table}

The displayed gains from iteration 10 to 50 account for approximately
84--91\% of the reported improvement from iteration 10 to 200.
Metric-selection differences prevent interpreting these shares as
protocol-harmonized optimization rates or a fitted logarithmic law.

\subsection{Calibration response by iteration count}

\begin{table}[H]
\centering
\small
\caption{SR-V2 ten-task mean accuracy gains with FragCal relative to Pile-128 (pp),
averaged over four seeds at each iteration count. Trend summarizes the
displayed sequence.}
\label{tab:iter_sensitivity}
\begin{tabular}{@{}lccccc@{}}
\toprule
Model & $i{=}10$ & $i{=}50$ & $i{=}100$ & $i{=}200$ & Trend \\
\midrule
Q3-1.7B  & +0.2 & +1.8 & +2.6 & +2.9 & $\nearrow$ monotonic \\
Q2.5-7B  & +0.6 & +0.0 & $-$0.2 & +0.4 & $\leftrightarrow$ flat \\
Ll2-7B   & +2.2 & +1.5 & +1.8 & +0.9 & $\searrow$ shrinks \\
Mi-7B    & $-$3.5 & +2.6 & +2.8 & +3.0 & $\nearrow$ grows \\
Q3-8B    & +1.8 & +1.2 & +1.6 & +1.6 & $\leftrightarrow$ flat \\
Phi-3    & $-$2.0 & +2.7 & +2.7 & +3.4 & $\nearrow$ grows \\
\bottomrule
\end{tabular}
\end{table}

Calibration response varies with iteration count and model. It grows for
some models, shrinks for Llama-2-7B, and is small for Qwen2.5-7B-Instruct.
These within-SR-V2 comparisons do not interpolate continuously between the
different single-pass and iterative algorithms.

\subsection{Accuracy-gain share at 50 iterations}

\begin{table}[H]
\centering
\small
\caption{Pile-128 accuracy gains between iteration counts (fractions).
The final column is the share of the total gain from iteration 10 to 200 reached
by step 50, $100\,\Delta_{10\to50}/\Delta_{10\to200}$.}
\label{tab:iter_convergence}
\begin{tabular}{@{}lccc@{}}
\toprule
Model & $10{\to}50$ gain & $50{\to}200$ gain & Gain share at $i{=}50$ (\%) \\
\midrule
Q3-1.7B  & +.117 & +.022 & 84\% \\
Q2.5-7B  & +.247 & +.035 & 88\% \\
Ll2-7B   & +.129 & +.024 & 84\% \\
Mi-7B    & +.113 & +.020 & 85\% \\
Q3-8B    & +.243 & +.023 & 91\% \\
Phi-3    & +.154 & +.027 & 85\% \\
\bottomrule
\end{tabular}
\end{table}

The sweep summaries are separate from the primary paired-calibration panel.
Their differences are computed before display rounding.

\section{Responses of instruction-tuned variants}
\label{app:instruction_tuning}

Qwen2.5-7B-Instruct and Phi-3-mini are instruction-tuned members of the primary
panel. Qwen2.5-7B-Instruct has the smallest primary SR-V2 pp response.
Here we compare FragCal with generic calibration on Mistral-Instruct and
Llama-2-chat, with their base counterparts as context.

\begin{table}[H]
\centering
\small
\caption{SR-V2 W2G128 ten-task mean accuracies (fractions).
$\Delta$ is the accuracy gain with FragCal relative to Pile-128 (pp).
GC is the percentage of the gap to full-precision accuracy recovered by FragCal.
Deltas precede display rounding.}
\begin{minipage}{\linewidth}
\centering
\begin{tabular}{@{}llccccc@{}}
\toprule
Model & Type & Pile-128$^\dagger$ & FragCal & FP Ref & $\Delta$ (pp) & GC \\
\midrule
Mistral-7B            & Base    & .5603 & .5888 & .6623 & +2.85 & +27.9\% \\
\textbf{Mistral-7B-Instruct} & \textbf{Instruct} & \textbf{.5650} & \textbf{.5817} & \textbf{.6397} & \textbf{+1.68} & \textbf{+22.5\%} \\
Llama-2-7B             & Base    & .5263 & .5356 & .6080 & +0.93 & +11.4\% \\
\textbf{Llama-2-7b-chat}     & \textbf{Instruct} & \textbf{.5382} & \textbf{.5536} & \textbf{.6073} & \textbf{+1.54} & \textbf{+22.3\%} \\
\bottomrule
\end{tabular}
\\[2pt]
{\footnotesize $^\dagger$Pile-128 control throughout this table. Separate
Pile-200 reference means are .5657 for Mistral-Instruct and .5356 for
Llama-2-chat. They are not used in the displayed deltas.}
\end{minipage}
\end{table}

Both additional instruction-tuned variants show positive responses.
These comparisons do not estimate a general instruction-tuning effect. Gap closure
also depends on each model's remaining full-precision gap.

\section{Transfer to GSM8K}
\label{app:gsm8k}

GSM8K~\citep{cobbe2021gsm8k} (8-shot, exact-match) evaluates generative
mathematical reasoning. It tests whether calibration gains extend beyond
the primary task suite. This benchmark is not used in recipe construction
or in the primary ten-task panel.

The evaluation contains 200 method/model/width/calibration/seed cells.
Pile-128 is available only for SR-V2. The forty corresponding RTN-opt
cells were not evaluated. Pile-200 and FragCal-DC have complete paired
coverage in this five-model, two-width panel.

\subsection{Cross-calibration means}

\begin{table}[H]
\centering
\small
\caption{GSM8K 8-shot strict exact-match accuracy (fraction),
averaged across five models and four seeds per calibration.
Dashes denote unevaluated controls.}
\begin{tabular}{@{}llccc@{}}
\toprule
Method & Bit & Pile-128 & Pile-200 & FragCal-DC \\
\midrule
\spclass~RTN-opt & W2G128 & --- & .0006 & .0008 \\
\spclass~RTN-opt & W3G128 & --- & .098 & .097 \\
\midrule
\itclass~SR-V2 & W2G128 & .056 & .058 & .048 \\
\itclass~SR-V2 & W3G128 & .139 & .135 & .142 \\
\bottomrule
\end{tabular}
\end{table}

\paragraph{Two-bit accuracy.}
At W2, RTN-opt is near zero on GSM8K while SR-V2 scores 0.048--0.058
depending on calibration.

\paragraph{Three-bit accuracy.}
At W3, SR-V2 remains above RTN-opt under both shared calibration conditions.
SR-V2's means range from 0.135 to 0.142 across calibrations. These are
descriptive model-panel means, not an equivalence test of calibration
effects.

\paragraph{Calibration transfer.}
At W2, SR-V2 with FragCal-DC is below Pile-200 on GSM8K (0.048 vs 0.058).
Thus the main-panel improvement does not extend to this task. The results
cover SR-V2 and RTN-opt only and neither establish a multi-method transfer
effect nor rule out item overlap in the primary task suite.

\section{Volume control and WikiText-2 perplexity}
\label{app:volume}
\label{app:wikitext}

We first examine whether calibration volume can account for the accuracy
response. We then report held-out language-modeling scores and their
relation to task accuracy (\cref{sec:loss_paradox}).

\subsection{Volume control}

FragCal writes 208 documents, compared with 128 or 200 for Pile controls.
The primary six-model SR-V2 comparison and the five-model task analysis
use Pile-200 (\cref{tab:sensitivity,tab:primary_models}).
Both show positive average response, but document
counts and token budgets are not equalized. The small accuracy differences
in \cref{tab:historical_volume} cannot eliminate this confound without
equivalent reference scores and seed pairings across the volume and FragCal
comparisons.

\begin{table}[H]
\centering\small
\caption{SR-V2 ten-task accuracy changes from Pile-128 to Pile-200 (pp),
shown per model and as a five-model mean.
Reference scores and seed pairings are not verified as matching the primary FragCal panel.}
\label{tab:historical_volume}
\begin{tabular}{@{}lrrr@{}}
\toprule
Model & W2G128 & W3G128 & W4G128\\
\midrule
\inputrows{generated/historical_volume_rows.tex}
\bottomrule
\end{tabular}
\end{table}

\paragraph{Phi-3 perplexity volume control.}
The recorded four-seed WikiText-2~\citep{merity2017wikitext} means for
Pile-128 versus Pile-200 are 20.10 versus 20.03 at W2, 11.19 versus 11.17
at W3, and 9.90 versus 9.84 at W4. These small perplexity differences on one
model are distinct from ten-task accuracy effects.

\subsection{WikiText-2 perplexity}

\subsubsection{SignRound V2 five-model panel}

\begin{table}[H]
\centering
\small
\caption{SR-V2 WikiText-2 perplexity (PPL, lower is better), G128.
$\Delta=$ Pile-200 minus FragCal in perplexity units (positive is better).
FP Ref is unquantized. Deltas precede display rounding.}
\begin{tabular}{@{}lcrrrrc@{}}
\toprule
& & \multicolumn{3}{c}{W2G128} & \multicolumn{2}{c}{W3G128}\\
\cmidrule(lr){3-5}\cmidrule(lr){6-7}
Model & FP Ref & Pile-200 & FragCal & $\Delta$ & Pile-200 & $\Delta$ \\
\midrule
Qwen3-1.7B & 25.3 & 70.9 & 63.4 & \textbf{+7.6} & 31.7 & +0.9 \\
Llama-2-7B & 7.8 & 12.4 & 12.4 & $-$0.0 & 8.6 & +0.1 \\
Mistral-7B & 7.6 & 13.1 & 12.5 & \textbf{+0.6} & 8.3 & +0.1 \\
Qwen2.5-7B-Instruct & 11.1 & 23.1 & 23.0 & +0.1 & 12.9 & $-$0.1 \\
Qwen3-8B & 14.6 & 29.8 & 28.7 & \textbf{+1.1} & 16.5 & $-$0.0 \\
\bottomrule
\end{tabular}
\end{table}

Three W2 models have visibly lower perplexity with FragCal. Differences on
the remaining models are small at the displayed precision. W3 changes are
smaller.

\subsubsection{Multi-method panel}

\begin{table}[H]
\centering
\small
\caption{Multi-method WikiText-2 perplexity (PPL, lower is better).
$\Delta=$ Pile minus FragCal in perplexity units. FP Ref is unquantized.
Scalar methods use G128. AQLM uses its W2-equivalent codebook.}
\begin{tabular}{@{}llccrr@{}}
\toprule
Method & Bit & Model & FP Ref & Pile & $\Delta$ PPL \\
\midrule
\spclass~AWQ & W3G128 & Qwen3-1.7B & 25.3 & 34.2 & $-$0.5 \\
\spclass~AWQ & W3G128 & Llama-2-7B & 7.8 & 9.0 & $-$0.0 \\
\spclass~AWQ & W3G128 & Mistral-7B & 7.6 & 8.7 & $-$0.0 \\
\spclass~AWQ & W3G128 & Qwen2.5-7B-Instruct & 11.1 & 13.4 & $-$0.1 \\
\spclass~AWQ & W3G128 & Qwen3-8B & 14.6 & 16.9 & +0.2 \\
\midrule
\spclass~GPTQ & W3G128 & Qwen3-1.7B & 25.3 & 53.0 & \textbf{$-$18.5} \\
\spclass~GPTQ & W3G128 & Llama-2-7B & 7.8 & 9.4 & $-$0.2 \\
\spclass~GPTQ & W3G128 & Mistral-7B & 7.6 & 9.7 & $-$0.3 \\
\spclass~GPTQ & W3G128 & Qwen2.5-7B-Instruct & 11.1 & 14.3 & $-$0.1 \\
\spclass~GPTQ & W3G128 & Qwen3-8B & 14.6 & 18.9 & +1.6 \\
\midrule
\itclass~OmniQuant & W3G128 & Qwen3-1.7B & 25.3 & 37.2 & +1.1 \\
\itclass~OmniQuant & W3G128 & Llama-2-7B & 7.8 & 9.0 & +0.1 \\
\itclass~OmniQuant & W3G128 & Mistral-7B & 7.6 & 9.0 & +0.0 \\
\itclass~OmniQuant & W3G128 & Qwen2.5-7B-Instruct & 11.1 & 13.5 & $-$0.1 \\
\itclass~OmniQuant & W3G128 & Qwen3-8B & 14.6 & 18.1 & $-$0.1 \\
\midrule
\itclass~AQLM & W2-equiv. & Qwen3-1.7B & 25.3 & 31.4 & $-$0.2 \\
\itclass~AQLM & W2-equiv. & Llama-2-7B & 7.8 & 9.6 & +0.1 \\
\itclass~AQLM & W2-equiv. & Mistral-7B & 7.6 & 9.2 & +0.1 \\
\itclass~AQLM & W2-equiv. & Qwen2.5-7B-Instruct & 11.1 & 28.7 & \textbf{+2.1} \\
\itclass~AQLM & W2-equiv. & Qwen3-8B & 14.6 & 17.4 & $-$0.2 \\
\bottomrule
\end{tabular}
\end{table}

GPTQ Qwen3-1.7B W3 worsens by 18.5 perplexity units, whereas GPTQ Qwen3-8B
improves by 1.6. The panel mixes widths, including OmniQuant at W3 rather
than its W2 designated comparison, so it does not independently replicate
a class effect. These heterogeneous outcomes do not identify a Hessian or
format-alignment cause.

\subsection{Held-out cross-entropy as a candidate proxy}
\label{app:wikitext_proxy}

The SR-V2 W2G128 proxy comparison pairs Pile-200 and FragCal across five
models and four calibration seeds. Both checkpoints in each pair are
evaluated on the same WikiText input at sequence length 2048.
Positive cross-entropy reduction means lower loss with FragCal. Positive
accuracy change means higher ten-task AVG.

\begin{table}[H]
\centering\small
\caption{SR-V2 W2G128 WikiText proxy and downstream response, averaged
over four paired seeds per model against Pile-200. Cross-entropy reduction is Pile minus FragCal in
nats per token. Accuracy change is FragCal minus Pile in pp.}
\label{tab:wikitext_proxy}
\begin{tabular}{@{}lrr@{}}
\toprule
Model & Cross-entropy reduction & Accuracy change (pp)\\
\midrule
\inputrows{generated/wikitext_proxy_rows.tex}
\bottomrule
\end{tabular}
\end{table}

The mean cross-entropy reduction is $0.03914$ nats per token, positive in
four of five model means. Its hierarchical-bootstrap 95\% interval is
$[-0.00115,+0.08759]$. We use 20,000 draws, resampling models and then seeds
within sampled models, with random seed 20260727.
Across the five model means, Pearson $r=0.778$ with exact two-sided
permutation $p=0.083$ over all $5!$ assignments of the accuracy means.
The descriptive 20-cell correlation is $r=0.234$. Those cells are not
20 independent model observations. With an interval spanning zero and only
five models, this is a candidate proxy, not a validated predictor or a
causal explanation of recovery.

\section{External-benchmark coverage}
\label{app:orthogonal}

We examine transfer to graduate-level science questions with
GPQA-Diamond~\citep{rein2023gpqa}, instruction following with
IFEval~\citep{zhou2023ifeval}, and challenging reasoning tasks with
BIG-Bench Hard (BBH)~\citep{suzgun2023bbh}.
The panel covers AWQ, GPTQ, OmniQuant, and AQLM. SR-V2 is not included.
BBH is retained only for Llama-2-7B and Mistral-7B because the
other tested reference models returned zero under the exact-match parser.
This filtering limits the interpretation of the pooled panel.

\begin{table}[H]
\centering\small
\caption{Pooled external-benchmark accuracy changes (fractions).
Counts are paired evaluation cells across heterogeneous
models, widths, tasks, and seeds, not independent models.}
\label{tab:orthogonal}
\begin{tabular}{@{}lrc@{}}
\toprule
Method & Mean FragCal minus Pile-200 & Paired cells\\
\midrule
\spclass~AWQ & $+0.008$ & 48\\
\spclass~GPTQ & $+0.002$ & 72\\
\midrule
\itclass~OmniQuant & $-0.005$ & 62\\
\itclass~AQLM & $-0.002$ & 20\\
\bottomrule
\end{tabular}
\end{table}

Mean responses have mixed signs across the four methods.
The 202 paired cells are not independent model replicates, and the
heterogeneous pooling does not support model-level class inference.
These descriptive means do not test calibration equivalence, measure
SR-V2's response, or resolve item overlap in the main task suite.

\section{Llama-3.1-70B per-task detail}
\label{app:70b_detail}

These tables expand the 70B comparison in \cref{sec:scale_verification}
by task and seed. Four-seed W2 results precede the single-seed extensions.
All task scores are accuracy fractions. AVG is their unweighted mean.
ARC-C/E denote ARC-Challenge/Easy, OBQA is OpenBookQA, and TQA MC1 is
TruthfulQA MC1. Summary spreads are reported seed standard deviations.

\subsection{SR-V2 at two bits}

\begin{table}[H]
\centering
\small
\caption{Llama-3.1-70B SR-V2 W2G128 per-task accuracies (fractions).
Columns are calibration seeds. Pile-128 has only seed 42.
AVG is the ten-task mean. The final row reports seed mean $\pm$ SD.}
\begin{tabular}{@{}lrrrrrrrrr@{}}
\toprule
& \multicolumn{4}{c}{Pile-200} & \multicolumn{4}{c}{FragCal} & Pile-128\\
\cmidrule(lr){2-5}\cmidrule(lr){6-9}
Task & s42 & s123 & s456 & s789 & s42 & s123 & s456 & s789 & s42\\
\midrule
ARC-C & .543 & .550 & .543 & .524 & .562 & .567 & .558 & .550 & .537 \\
ARC-E & .813 & .817 & .813 & .814 & .810 & .819 & .802 & .815 & .808 \\
BoolQ & .819 & .845 & .850 & .821 & .836 & .831 & .831 & .831 & .841 \\
HellaSwag & .770 & .773 & .772 & .768 & .790 & .792 & .795 & .787 & .762 \\
LAMBADA & .760 & .753 & .758 & .750 & .749 & .749 & .761 & .753 & .764 \\
MMLU & .643 & .645 & .643 & .644 & .676 & .680 & .669 & .668 & .626 \\
OBQA & .432 & .440 & .440 & .446 & .448 & .464 & .448 & .446 & .424 \\
PIQA & .783 & .794 & .791 & .785 & .797 & .801 & .800 & .791 & .784 \\
TQA MC1 & .245 & .253 & .246 & .242 & .238 & .257 & .230 & .251 & .238 \\
WinoGrande & .767 & .749 & .754 & .766 & .759 & .772 & .757 & .766 & .754 \\
AVG & .657 & .662 & .661 & .656 & .667 & .673 & .665 & .666 & .654 \\
\midrule
Mean AVG & \multicolumn{4}{c}{.659 $\pm$ .003} & \multicolumn{4}{c}{.668 $\pm$ .004} & ---\\
\bottomrule
\end{tabular}
\end{table}

SR-V2's FragCal$-$Pile-200 response is $\Delta = +0.86$ pp over four seeds
(FragCal $=0.6676\pm0.0037$ versus Pile-200 $=0.6590\pm0.0029$).
The single Pile-128 score of 0.654 is a separate reference, not a four-seed
volume-control experiment.

MMLU has the largest SR-V2 task gain at 70B, consistent with the 7B pattern.
It gains +2.95 pp over four seeds and +3.24 pp at seed 42.

\subsection{References and other two-bit methods}

\begin{table}[H]
\centering
\small
\caption{Llama-3.1-70B unquantized BF16 and data-free RTN W2G128
reference accuracies (fractions, one evaluation each).
Tasks use 0-shot evaluation except MMLU (5-shot). AVG is the ten-task mean.}
\label{tab:70b_ref}
\begin{tabular}{@{}lrr@{}}
\toprule
Task & BF16 Ref & RTN W2G128\\
\midrule
ARC-C & .613 & .235 \\
ARC-E & .851 & .285 \\
BoolQ & .872 & .400 \\
HellaSwag & .857 & .264 \\
LAMBADA & .793 & .003 \\
MMLU & .763 & .231 \\
OBQA & .480 & .264 \\
PIQA & .826 & .545 \\
TQA MC1 & .341 & .242 \\
WinoGrande & .818 & .503 \\
AVG & \textbf{.7215} & \textbf{.2972} \\
\bottomrule
\end{tabular}
\end{table}

\begin{table}[H]
\centering
\small
\caption{Llama-3.1-70B RTN-opt W2G128 per-task accuracies (fractions).
RTN-opt uses scale search without iterative weight tuning.
AVG is the ten-task mean. The final row gives seed mean $\pm$ SD.}
\label{tab:70b_rtnopt}
\begin{tabular}{@{}lrrrrrrrr@{}}
\toprule
& \multicolumn{4}{c}{Pile-200} & \multicolumn{4}{c}{FragCal}\\
\cmidrule(lr){2-5}\cmidrule(lr){6-9}
Task & s42 & s123 & s456 & s789 & s42 & s123 & s456 & s789\\
\midrule
ARC-C & .220 & .213 & .230 & .221 & .219 & .224 & .227 & .226 \\
ARC-E & .334 & .341 & .344 & .327 & .348 & .343 & .344 & .342 \\
BoolQ & .502 & .498 & .500 & .494 & .498 & .515 & .503 & .513 \\
HellaSwag & .283 & .276 & .281 & .280 & .282 & .281 & .282 & .279 \\
LAMBADA & .054 & .057 & .060 & .054 & .058 & .056 & .057 & .057 \\
MMLU & .233 & .229 & .233 & .231 & .238 & .237 & .236 & .237 \\
OBQA & .250 & .254 & .252 & .262 & .256 & .264 & .260 & .270 \\
PIQA & .567 & .569 & .574 & .563 & .565 & .566 & .569 & .558 \\
TQA MC1 & .261 & .262 & .263 & .257 & .252 & .260 & .267 & .263 \\
WinoGrande & .511 & .507 & .500 & .504 & .498 & .504 & .505 & .499 \\
AVG & .322 & .321 & .324 & .319 & .322 & .325 & .325 & .324 \\
\midrule
Mean AVG & \multicolumn{4}{c}{.3213 $\pm$ .0019} & \multicolumn{4}{c}{.3239 $\pm$ .0016}\\
\bottomrule
\end{tabular}
\end{table}

\begin{table}[H]
\centering
\small
\caption{Llama-3.1-70B GPTQ W2G128 per-task accuracies (fractions).
AVG is the ten-task mean. The final row gives seed mean $\pm$ SD.
FragCal improves accuracy by $+2.33$ pp relative to Pile, recovering $6.6\%$ of the BF16 gap.}
\label{tab:70b_gptq}
\begin{tabular}{@{}lrrrrrrrr@{}}
\toprule
& \multicolumn{4}{c}{Pile-200} & \multicolumn{4}{c}{FragCal}\\
\cmidrule(lr){2-5}\cmidrule(lr){6-9}
Task & s42 & s123 & s456 & s789 & s42 & s123 & s456 & s789\\
\midrule
ARC-C & .271 & .267 & .273 & .257 & .311 & .276 & .285 & .271 \\
ARC-E & .392 & .399 & .435 & .412 & .460 & .476 & .509 & .444 \\
BoolQ & .534 & .562 & .536 & .520 & .502 & .497 & .557 & .562 \\
HellaSwag & .405 & .428 & .427 & .390 & .481 & .466 & .481 & .446 \\
LAMBADA & .147 & .166 & .141 & .137 & .201 & .205 & .196 & .146 \\
MMLU & .234 & .234 & .238 & .237 & .260 & .251 & .263 & .248 \\
OBQA & .264 & .266 & .284 & .244 & .274 & .272 & .296 & .282 \\
PIQA & .605 & .615 & .610 & .622 & .639 & .640 & .652 & .627 \\
TQA MC1 & .266 & .241 & .257 & .247 & .256 & .239 & .263 & .241 \\
WinoGrande & .519 & .513 & .539 & .527 & .544 & .528 & .537 & .514 \\
AVG & .364 & .369 & .374 & .359 & .393 & .385 & .404 & .378 \\
\midrule
Mean AVG & \multicolumn{4}{c}{.3665 $\pm$ .0064} & \multicolumn{4}{c}{.3899 $\pm$ .0110}\\
\bottomrule
\end{tabular}
\end{table}

The FragCal-calibrated method means are approximately 0.324 for RTN-opt,
0.390 for GPTQ, and 0.668 for SR-V2, versus 0.297 for data-free RTN and
0.7215 for the unquantized reference. These are different methods, not
nested interventions isolating an additive contribution from each component.

The GPTQ W2 configuration is above the operational 0.35 threshold.
Its positive response does not imply near-full-precision recovery.

\subsection{Single-seed extensions}
Additional seed-42 cells cover AWQ at W2 and W3, GPTQ at W3, and SR-V2 at W3.
They are reported as point estimates without a seed-variability estimate.

\paragraph{AWQ at two bits.}
AWQ gives AVG 0.322 with Pile and 0.321 with FragCal, a $-0.11$ pp response
at seed 42, with zero LAMBADA accuracy. The GPTQ W2 configuration instead
has a functional baseline and a positive response.

\paragraph{Three-bit responses.}
The W3 panel compares AWQ, GPTQ, and SR-V2 at seed 42.

\begin{table}[H]
\centering
\small
\caption{Llama-3.1-70B W3G128 single-seed (s42) accuracies (fractions).
Pile denotes Pile-200. $\Delta$ is FragCal minus Pile (pp), calculated
before rounding. AVG is the ten-task mean.}
\label{tab:70b_above_cliff_w3}
\begin{tabular}{@{}lrrrrrrrrr@{}}
\toprule
& \multicolumn{3}{c}{\spclass~AWQ} & \multicolumn{3}{c}{\spclass~GPTQ} & \multicolumn{3}{c}{\itclass~SR-V2}\\
\cmidrule(lr){2-4}\cmidrule(lr){5-7}\cmidrule(lr){8-10}
Task & Pile & FragCal & $\Delta$ & Pile & FragCal & $\Delta$ & Pile & FragCal & $\Delta$\\
\midrule
ARC-C & .584 & .597 & $+1.28$ & .573 & .585 & $+1.28$ & .603 & .624 & $+2.05$\\
ARC-E & .824 & .834 & $+0.93$ & .815 & .819 & $+0.38$ & .838 & .843 & $+0.50$\\
BoolQ & .850 & .847 & $-0.34$ & .823 & .815 & $-0.76$ & .864 & .852 & $-1.16$\\
HellaSwag & .832 & .833 & $+0.02$ & .818 & .815 & $-0.31$ & .827 & .837 & $+0.99$\\
LAMBADA & .768 & .763 & $-0.54$ & .755 & .748 & $-0.60$ & .790 & .787 & $-0.29$\\
MMLU & .728 & .736 & $+0.86$ & .685 & .684 & $-0.18$ & .741 & .746 & $+0.54$\\
OBQA & .476 & .474 & $-0.20$ & .436 & .452 & $+1.60$ & .464 & .474 & $+1.00$\\
PIQA & .820 & .816 & $-0.49$ & .829 & .823 & $-0.59$ & .821 & .824 & $+0.32$\\
TQA MC1 & .295 & .291 & $-0.34$ & .294 & .308 & $+1.36$ & .322 & .320 & $-0.24$\\
WinoGrande & .796 & .804 & $+0.79$ & .777 & .766 & $-1.10$ & .806 & .800 & $-0.55$\\
AVG & .6975 & .6994 & $+0.19$ & .6804 & .6815 & $+0.12$ & .7074 & .7106 & $+0.32$\\
\bottomrule
\end{tabular}
\end{table}

All three Pile W3 baselines exceed the 0.62 upper descriptive guide,
and all three calibration responses have magnitude below 0.5 pp. These
point estimates do not establish statistical equivalence. Together with the W2
cases, these results show that the same model can occupy different accuracy
regimes across quantizers and widths. They do not constitute an across-model replication of the class
contrast.

\section{Comparison with published SignRound V2 results}
\label{app:srv2_repro}

We compare our Llama-2-7B W2G128 evaluations with Table 8 of SignRound V2
version 1~\citep{cheng2025signroundv2}. Both settings use Pile calibration,
128 samples, group size 128, and sequence length 2048. Matching
optimization batch sizes is not established.
Our headline AVG uses ten tasks, whereas that publication reports a
five-task mean over ARC-C, ARC-E, HellaSwag, PIQA, and WinoGrande.
This Pile-128 comparison is separate from the Pile-200/FragCal
configuration inventory in \cref{app:implementation}.

The pipelines also differ in HellaSwag scoring and implementation. The
published comparison uses raw HellaSwag accuracy. Our evaluations
use length-normalized accuracy. The AutoRound implementation used in the
study~\citep{autoround_github} allows a per-group signed-scale choice between
two two-bit grids. The published v1 implementation and the evaluated
LWC-only OmniQuant configuration are not fully matched controls.

\begin{table}[H]
\centering\small
\caption{Llama-2-7B W2G128 task-mean accuracies (\%). $\Delta$ is Ours
minus Pub.\ (pp), calculated before rounding.
Four-task means exclude HellaSwag (HS). Implementation differences remain.
``Pub.'' refers to SignRound V2 version 1.}
\label{tab:srv2_repro_5task}
\begin{tabular}{@{}lrrrrrr@{}}
\toprule
 & \multicolumn{3}{c}{Five tasks} & \multicolumn{3}{c}{Four tasks, excluding HS}\\
\cmidrule(lr){2-4}\cmidrule(lr){5-7}
Method & Ours & Pub. & $\Delta$ & Ours & Pub. & $\Delta$\\
\midrule
Unquantized & 68.59 & 64.66 & $+3.94$ & 66.76 & 66.50 & $+0.26$\\
\spclass~AWQ & 35.89 & 34.74 & $+1.15$ & 38.34 & 37.00 & $+1.34$\\
\spclass~GPTQ & 39.08 & 41.56 & $-2.47$ & 40.76 & 43.80 & $-3.04$\\
\midrule
\itclass~OmniQuant & 45.22 & 46.98 & $-1.76$ & 45.51 & 48.65 & $-3.14$\\
\itclass~SR-V2 & 64.73 & 57.88 & $+6.85$ & 63.51 & 60.30 & $+3.21$\\
\bottomrule
\end{tabular}
\end{table}

The four-task SR-V2 gap of $+3.21$ pp
cannot be attributed entirely to a codebook change without an isolated
implementation ablation. FragCal response is measured within
a fixed implementation and reference condition. Comparing absolute scores
across papers also requires matching those conditions.

\section{Hessian changes with calibration}
\label{app:hessian}

To test whether the small dense-panel GPTQ response coincides with similar
calibration statistics, we compute the input second-moment matrix
$\mathbf{H} = \frac{1}{n}\sum_{i=1}^{n} \mathbf{X}_i \mathbf{X}_i^\top$
from Pile-128 and FragCal at seed 42. The spot check covers three models,
three layers per model (early, middle, final), and two modules per layer,
the query projection (\texttt{q\_proj}) and feed-forward gate projection
(\texttt{gate\_proj}), for eighteen comparisons.

\begin{table}[H]
\centering
\small
\caption{Hessian changes from Pile to FragCal.
$\|\Delta H\|_F / \|H_{\text{Pile}}\|_F$ is relative Frobenius divergence.
$\cos\theta_1$ and $\cos\theta_3$ are cosine similarities of the 1st and 3rd eigenvectors.}
\label{tab:hessian}
\begin{tabular}{@{}lllccc@{}}
\toprule
Model & Layer & Module & $\|\Delta H\|_F / \|H_{\text{Pile}}\|_F$ & $\cos\theta_1$ & $\cos\theta_3$ \\
\midrule
Qwen3-1.7B  & 0 (early)  & gate\_proj & 15.2\% & 0.995 & 0.943 \\
Qwen3-1.7B  & 0 (early)  & q\_proj   & 14.8\% & 0.988 & 0.983 \\
Qwen3-1.7B  & 13 (mid)   & gate\_proj & 28.2\% & 0.980 & 0.249 \\
Qwen3-1.7B  & 13 (mid)   & q\_proj   & 16.6\% & 0.994 & 0.489 \\
Qwen3-1.7B  & 27 (final) & gate\_proj & 23.0\% & 0.987 & 0.896 \\
Qwen3-1.7B  & 27 (final) & q\_proj   & 30.4\% & 0.977 & 0.112 \\
\midrule
Llama-2-7B  & 0 (early)  & gate\_proj & 21.6\% & 0.989 & 0.337 \\
Llama-2-7B  & 0 (early)  & q\_proj   &  7.0\% & 0.999 & 0.991 \\
Llama-2-7B  & 15 (mid)   & gate\_proj & 52.4\% & 0.931 & 0.437 \\
Llama-2-7B  & 15 (mid)   & q\_proj   & 23.7\% & 0.987 & 0.922 \\
Llama-2-7B  & 31 (final) & gate\_proj & 41.2\% & 0.956 & 0.159 \\
Llama-2-7B  & 31 (final) & q\_proj   & 41.8\% & 0.955 & 0.505 \\
\midrule
Qwen3-8B    & 0 (early)  & gate\_proj & 43.4\% & 0.936 & 0.386 \\
Qwen3-8B    & 0 (early)  & q\_proj   & 19.5\% & 0.992 & 0.914 \\
Qwen3-8B    & 18 (mid)   & gate\_proj & 33.4\% & 0.972 & 0.653 \\
Qwen3-8B    & 18 (mid)   & q\_proj   & 20.5\% & 0.988 & 0.973 \\
Qwen3-8B    & 35 (final) & gate\_proj & 26.0\% & 0.983 & 0.773 \\
Qwen3-8B    & 35 (final) & q\_proj   & 24.4\% & 0.985 & 0.657 \\
\midrule
\multicolumn{3}{@{}l}{\textbf{Mean (18 comparisons)}} & \textbf{26.8\%} & \textbf{0.977} & \textbf{0.632} \\
\bottomrule
\end{tabular}
\end{table}

Calibration changes the measured Hessians. Mean relative Frobenius divergence is 26.8\% (range 7.0--52.4\%).
The top eigenvector is well-preserved ($\cos\theta_1 > 0.93$ in 18/18 comparisons), but the 3rd eigenvector diverges substantially ($\cos\theta_3 < 0.5$ in 7/18 comparisons).
The largest change is at Llama-2-7B layer 15 \texttt{gate\_proj} (52.4\%).
These statistics do not imply identical GPTQ rounding decisions. GPTQ uses
a damped inverse-Hessian update, whose behavior is not determined by the
leading eigenvector alone. The downstream results in \cref{sec:results}
do not identify Hessian robustness as a universal mechanism.

\section{Mechanism probes and interpretation}
\label{app:mechanism}

We examine how group scales change with calibration and optimization,
and what happens when learned scales are flattened. These probes develop
the scale analysis in \cref{sec:mechanism}. Method class, quantizer
parameterization, and operating regime are not independently varied.

\subsection{Scale variation across models and widths}

For a fixed model, method, width, and seed, we flatten the saved per-group
scales within each linear module into a vector $s$. We define
\[
c(s)=\frac{\sigma(s)}{|\mu(s)|+10^{-12}}, \qquad
d_{\ell k}=\frac{|c(s_{\mathrm{FragCal},\ell k})-
                    c(s_{\mathrm{Pile},\ell k})|}
                   {|c(s_{\mathrm{Pile},\ell k})|+10^{-9}}.
\]
Here $\mu$ and $\sigma$ are the mean and population standard deviation
over scales within the module, not across seeds.
For each seed $k$, we average $d_{\ell k}$ over modules $\ell$ present in
both calibration arms. We then average these seed means equally over
seeds 42, 123, 456, and 789.
\Cref{tab:mechanism_3family} reports the ratio of these method means,
SR-V2 divided by GPTQ, not the mean of module-level method ratios.
Signed scales are not converted to magnitudes before computing $\mu$
or $\sigma$. The denominator uses the absolute mean, not the mean
absolute scale. Cancellation in signed scales, particularly for Phi-3,
limits comparability across scale conventions.

Ratios exceed one in all nine recorded cells, but the largest ratios for
Llama-2-7B and Qwen3-1.7B occur at W4 rather than at the width with the largest
accuracy improvement. The sensitivity ratio is therefore not a standalone
accuracy predictor.

\subsection{Optimization trajectories}
\label{app:c3_trajectory}

\begin{figure}[t]
\centering
\includegraphics[width=0.85\linewidth]{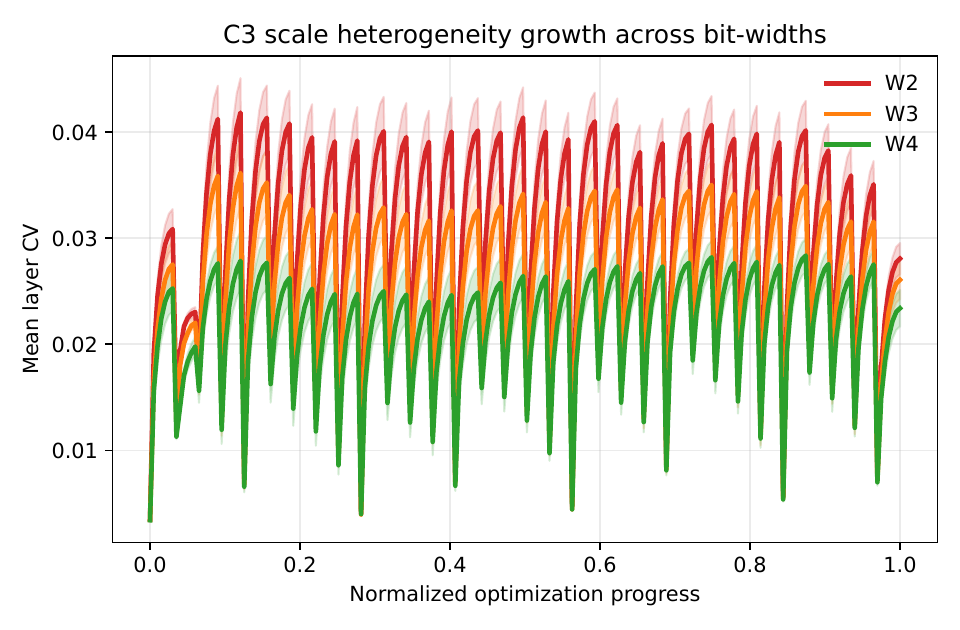}
\caption{Mean layer scale-CV trajectories across normalized
optimization progress. Bands show one standard deviation across four seeds
(variance divisor $n=4$).
CV grows at all three widths.}
\label{fig:c3_cv_progress}
\end{figure}

\begin{figure}[t]
\centering
\includegraphics[trim=0bp 9bp 552bp 24bp,clip,width=0.60\linewidth]{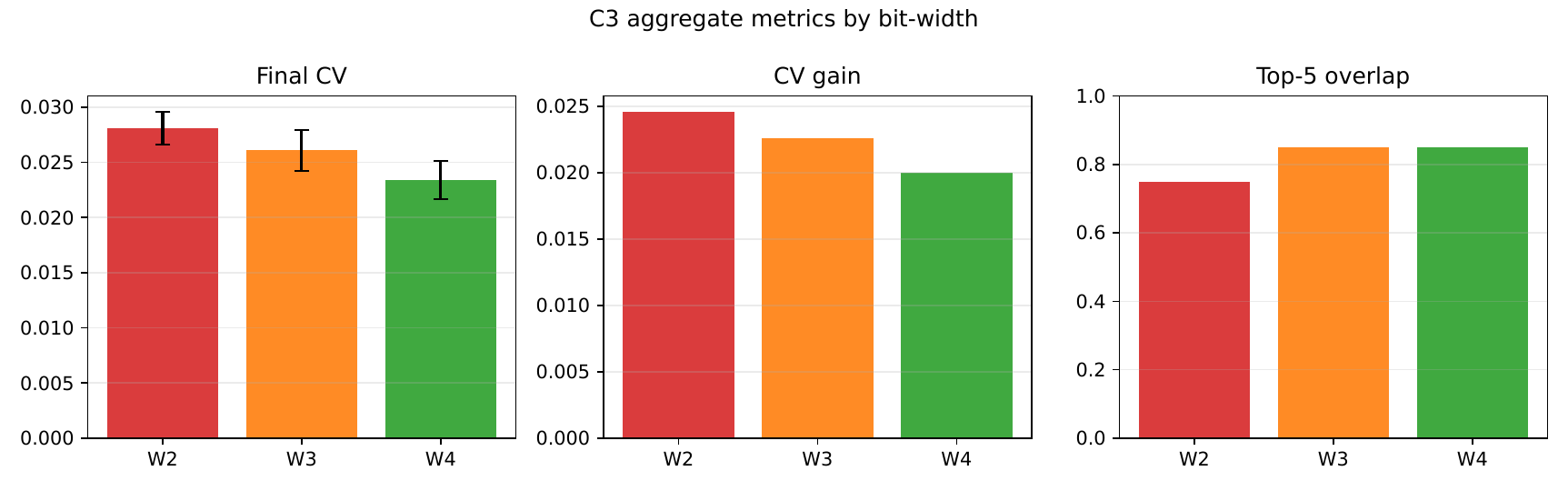}\\
\includegraphics[trim=276bp 9bp 276bp 24bp,clip,width=0.60\linewidth]{generated/fig_c3_aggregate_metrics.pdf}\\
\includegraphics[trim=552bp 9bp 0bp 24bp,clip,width=0.60\linewidth]{generated/fig_c3_aggregate_metrics.pdf}
\caption{Final CV, CV gain, and layer overlap by width (\cref{tab:c3_trajectory_summary}).
Final-CV error bars show one standard deviation across four seeds (variance divisor $n=4$).
Overlap compares the five highest-CV layers early and late in optimization.}
\label{fig:c3_aggregate_metrics}
\end{figure}

\begin{figure}[t]
\centering
\begin{minipage}{0.67\linewidth}
\includegraphics[trim=0bp 8bp 576.2bp 24bp,clip,width=\linewidth]{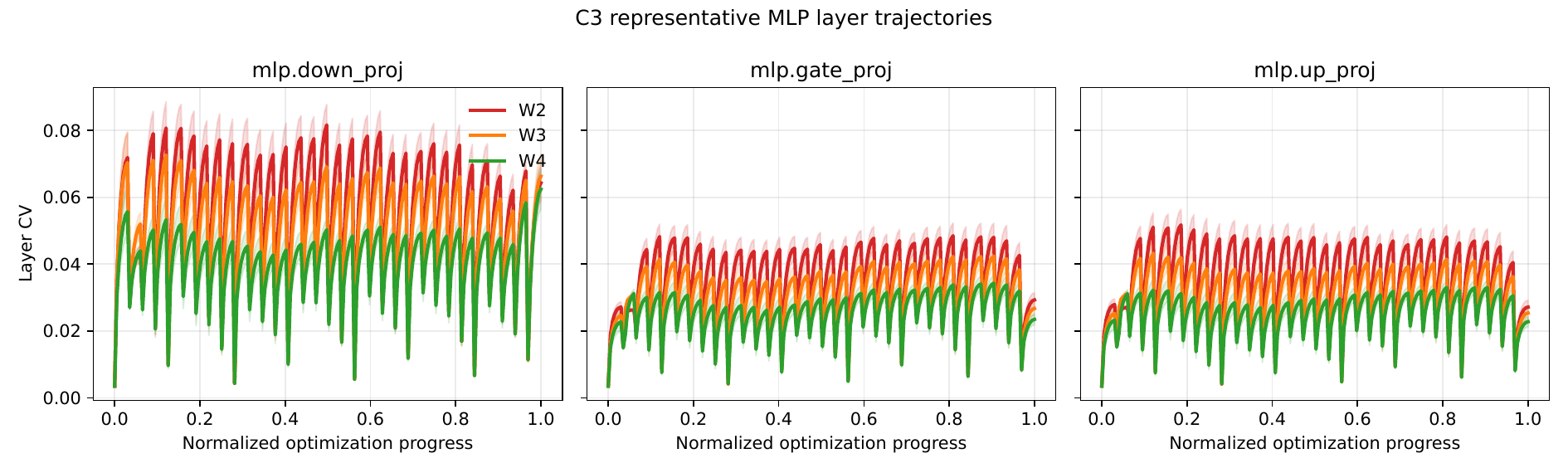}\\
\makebox[\linewidth][r]{\includegraphics[trim=331bp 8bp 290.64bp 24bp,clip,width=0.86272\linewidth]{generated/fig_c3_representative_layers.pdf}}\\
\makebox[\linewidth][r]{\includegraphics[trim=616.56bp 8bp 5.08bp 24bp,clip,width=0.86272\linewidth]{generated/fig_c3_representative_layers.pdf}}
\end{minipage}
\caption{Representative MLP-projection scale-CV trajectories, with
shared vertical scale and normalized optimization progress on the horizontal axes.
Bands show one standard deviation across four seeds (variance divisor $n=4$).
The displayed trajectories oscillate during optimization.}
\label{fig:c3_representative_layers}
\end{figure}

The SR-V2 scale-trajectory probe (labeled C3 in the plots) uses Llama-2-7B
with Pile-10k calibration and covers W2G128, W3G128, and W4G128 with
four seeds each. Plot legends abbreviate these
as W2, W3, and W4. \Cref{tab:c3_trajectory_summary} reports final
CV, net CV growth, and overlap between the five highest-CV layers early
and late in optimization. These values
describe scale statistics. They do not identify the layers' causal
importance for a task.
\Cref{fig:c3_cv_progress,fig:c3_aggregate_metrics,fig:c3_representative_layers}
show the mean trajectory, aggregate metrics, and representative projections,
respectively.

\begin{table}[H]
\centering
\small
\caption{Scale-trajectory summary over four seeds per width. Final CV and
CV growth are reported as mean $\pm$ standard deviation across seeds
(variance divisor $n=4$).
Overlap is the mean overlap between the five highest-CV layers early and late in optimization.}
\label{tab:c3_trajectory_summary}
\begin{tabular}{lcccc}
\toprule
Width & Runs & Final CV & $\Delta$CV & Top-5 overlap \\
\midrule
W2G128 & 4 & 0.0281 $\pm$ 0.0015 & 0.0246 $\pm$ 0.0015 & 0.75 \\
W3G128 & 4 & 0.0261 $\pm$ 0.0018 & 0.0226 $\pm$ 0.0018 & 0.85 \\
W4G128 & 4 & 0.0234 $\pm$ 0.0017 & 0.0200 $\pm$ 0.0017 & 0.85 \\
\bottomrule
\end{tabular}
\end{table}

\clearpage
\subsection{Fixed-code scale flattening}
\label{app:followup_knockout}

Starting from a FragCal-calibrated checkpoint, the probe replaces scales by
\[
s_\alpha[g,c]=\operatorname{sign}(s[g,c])
\left((1-\alpha)|s[g,c]|+
\frac{\alpha}{G}\sum_{g'=1}^{G}|s[g',c]|\right).
\]
The indices denote group and channel. The sum averages scale magnitudes
over all groups within that channel.
Integer codes and zero points remain fixed. At $\alpha=1$, within-channel
scale magnitudes are equalized, introducing a mismatch with the original
codes. Signed-scale conventions mean this operation need not remove every
form of signed-scale variation.

\begin{table}[H]
\centering\small
\caption{Llama-2-7B SR-V2 W3G128 fixed-code flattening. LAMBADA accuracy
and changes with FragCal relative to Pile are fractions, reported as four-seed
mean $\pm$ sample SD except for the labeled single-seed row.
Dashes denote unreported differences.}
\label{tab:followup_knockout}
\begin{tabular}{@{}lcc@{}}
\toprule
Condition & LAMBADA & FragCal minus Pile\\
\midrule
$\alpha=0$ & $0.7082\pm0.0034$ & $-0.0115\pm0.0037$\\
$\alpha=0.1$ & $0.7072\pm0.0040$ & $-0.0126\pm0.0042$\\
$\alpha=0.25$ & $0.6982\pm0.0028$ & ---\\
$\alpha=0.5$, seed 42 & 0.6499 & ---\\
$\alpha=1$ & $0.0000\pm0.0000$ & ---\\
Pile control & $0.7197\pm0.0042$ & ---\\
\bottomrule
\end{tabular}
\end{table}

The separate Llama-3.1-70B W2 probe reports an AVG decline from
$0.6683\pm0.0029$ to $0.2954\pm0.0009$ under full flattening.
Both probes show that post-hoc scale flattening damages fidelity when
codes remain fixed. Reoptimized-code compensation remains untested, so the
intervention does not establish scale variation as a sufficient explanation
of task-specific gains.

\subsection{Evidence map}
\label{app:hypothesis_table}

\begin{table}[H]
\centering\small
\caption{What the existing probes establish and leave unresolved.}
\label{tab:hypothesis_registry}
\begin{tabularx}{\linewidth}{@{}lXX@{}}
\toprule
Probe & Observation & Interpretation boundary\\
\midrule
Wrong designations & Most of the gain remains & Options and task exposure remain\\
RTN-opt & Small response versus SR-V2's gain & Iteration remains coupled to other method differences\\
Same W3 & Pairwise GC favors SR-V2 over AWQ & GC uses method-specific controls. Pile-200 pp class interval spans zero\\
70B & Both GPTQ and SR-V2 improve at W2 & Extends sensitivity. Scale and headroom remain coupled\\
Scale flattening & Fixed-code fidelity degrades & Code reoptimization remains untested\\
Hessian spot check & Calibration statistics differ & Rounding-decision changes are unmeasured\\
Fixed-reference loss & Error reduction and accuracy gain do not consistently coincide & Incomplete proxy in these comparisons\\
\bottomrule
\end{tabularx}
\end{table}

Distribution alignment and calibration-shaped optimization remain compatible
explanations.

\section{Recipe comparisons and extended evaluations}
\label{app:rebuttal}

These checks add recipe comparisons, transfer replication, and item-overlap
evidence to \cref{sec:recipe_controls,sec:overlap}. We also examine a
mixture-of-experts (MoE) model and joint weight-activation quantization
(\cref{sec:scale_verification}).

\subsection{Alternative calibration recipes}
\label{app:alternatives}

The SR-V2 W2 alternative-recipe panel uses four seeds and the five models
in the paired task analysis (\cref{tab:primary_models}).
Diversity-DC selects for diversity,
LSA-DC uses a latent-semantic selection criterion, and ArXiv-OOD supplies
scientific prose. The table uses same-harness rescoring of the calibrated
checkpoints for these comparisons.
These experiments do not include high-perplexity Pile selection or
activation-space diversity selection. They do not establish an optimal recipe.

\begin{table}[H]
\centering\small
\caption{SR-V2 W2G128 gains in ten-task mean accuracy with FragCal relative to
alternative calibration recipes (pp), averaged over four seeds per model. Intervals are
model-level 95\% $t$ intervals on five model means.}
\label{tab:alternatives}
\begin{tabular}{@{}lrrc@{}}
\toprule
Alternative & $\Delta$ pp & 95\% interval & Positive models\\
\midrule
\inputrows{generated/alternative_rows.tex}
\bottomrule
\end{tabular}
\end{table}

The independent OmniQuant LWC-only comparison covers Qwen3-1.7B,
Qwen2.5-7B-Instruct, and Phi-3. The FragCal advantages over
Diversity-DC, LSA-DC, and ArXiv-OOD are $+1.42$, $+1.01$, and $+2.65$ pp.
The hierarchical-bootstrap intervals are $[+0.67,+2.40]$,
$[+0.56,+1.62]$, and $[+2.17,+3.10]$, respectively. These intervals use a
different procedure from the SR-V2 model-level $t$ intervals.

\subsection{Unranked Random-QA control}
\label{app:random_qa}

Random QA draws 200 items uniformly from the combined MMLU, ARC/OpenBookQA,
and HellaSwag source pools, without family balancing or a Pile anchor.
Because the pools differ in size, the recorded mixtures contain
66--69\% MMLU, 26--28\% HellaSwag, and 5.2--6.3\% ARC/OpenBookQA.
FragCal instead retains generic text and balances its task supplements.
The comparison changes both source allocation and generic-text coverage.
It does not isolate the effect of fragility ranking.

\begin{table}[H]
\centering\small
\caption{SR-V2 W2G128 ten-task accuracy gains (pp), using four-seed model means.
References are Pile-128 for five models and Pile-200 for Phi-3.
Random QA uses 200 question-answer items. FragCal writes 208 documents including a Pile anchor.
This panel is separate from the five-model Pile-200 answer controls.}
\label{tab:random_qa}
\begin{tabular}{@{}lrr@{}}
\toprule
Model & Random QA minus Pile & FragCal minus Pile\\
\midrule
\inputrows{generated/random_qa_rows.tex}
\midrule
Six-model mean & $+1.46$ & $+1.93$\\
\bottomrule
\end{tabular}
\end{table}

Random QA improves five of six model means, but Qwen2.5-7B-Instruct loses
0.86 pp and Qwen3-8B is nearly unchanged. The recorded model-seed comparisons
are positive in 18/24 cases for Random QA and 24/24 for FragCal. FragCal has
the higher panel mean and fewer regressions, but Random QA gains more on
Llama-2 and Mistral. The advantage is model-dependent.

\subsection{Transfer from ARC/OBQA to MMLU}
\label{app:transfer}

This four-seed comparison extends the seed-42 result in
\cref{tab:cross_family}. Each arm adds 32 supplemental documents to Pile-128.
The arms differ in whether the supplement comes from ARC/OBQA or MMLU.
The outcome is MMLU accuracy under SR-V2 W2. Five models, two arms, and four
seeds give forty evaluations and twenty paired contrasts.

\begin{table}[H]
\centering\small
\caption{SR-V2 W2G128 MMLU accuracy gain with ARC/OBQA32 versus
MMLU32 supplements to Pile-128 (four paired seeds per model). Model-balanced
mean $+1.50$ pp, hierarchical-bootstrap 95\% interval $[+0.67,+2.46]$.
All five model means and 19/20 seed pairs are positive.}
\label{tab:transfer_replication}
\begin{tabular}{@{}lrc@{}}
\toprule
Model & Mean $\Delta$ pp & Positive seeds\\
\midrule
\inputrows{generated/transfer_rows.tex}
\bottomrule
\end{tabular}
\end{table}

This comparison is distinct from matching STEM and humanities domains
within MMLU, for which the available evidence is inconclusive. Shared QA
format, topics, and possible source overlap prevent a unique mechanistic
interpretation of the cross-family result.

\subsection{Item-overlap audit}
\label{app:overlap}

This check searches for content overlap across the named training and
evaluation splits.

\begin{table}[H]
\centering\small
\caption{Normalized held-out question-key/completion matches in the experimental FragCal-DC files.
These counts do not establish full question-plus-options duplication or quantify its accuracy effect.}
\label{tab:overlap}
\begin{tabular}{@{}crrrrr@{}}
\toprule
Seed & MMLU & ARC-C & ARC-E & OBQA & HellaSwag\\
\midrule
\inputrows{generated/overlap_rows.tex}
\bottomrule
\end{tabular}
\end{table}

The recorded question-key matches originate in the combined ARC/OBQA slot.
The two HellaSwag completion matches originate in the HellaSwag slot.
Some MMLU question keys are generic. Source-level full-item verification and
a matched-item exclusion sensitivity analysis are needed before attributing
an accuracy effect to these matches. Neither zero overlap nor a contamination
adjusted gain is claimed here.

\subsection{MoE and weight-activation extensions}
\label{app:extensions}

Mixtral-8x7B-Instruct~\citep{jiang2024mixtral} is evaluated at W3G128 on ARC-Challenge only. SR-V2
changes from 0.6305 to 0.6408, $+1.02$ pp with paired-seed SD 0.39 pp.
GPTQ changes from 0.6165 to 0.6026, $-1.39$ pp with SD 0.30 pp.
All four seeds agree in direction for each method. The BF16 score is
0.6502. This is a one-model, one-task result. Routing behavior was not measured.

\begin{table}[H]
\centering\small
\caption{Accuracy gains with FragCal relative to Pile-200 when quantizing
weights and activations with AutoRound (ten-task mean, pp).
Intervals are within-model paired-seed 95\% $t$ intervals over four seeds.}
\label{tab:wa_extensions}
\begin{tabular}{@{}llrc@{}}
\toprule
Model & Format & $\Delta$ pp & 95\% interval\\
\midrule
Phi-3 & W2A8G128 & $+3.00$ & $[+0.99,+5.01]$\\
Mistral-7B & W2A8G128 & $+2.25$ & $[+1.93,+2.57]$\\
Phi-3 & MXINT W2A8G32 & $+3.29$ & $[+1.44,+5.15]$\\
Mistral-7B & MXINT W2A8G32 & $+3.10$ & $[+1.59,+4.62]$\\
\bottomrule
\end{tabular}
\end{table}

MXINT uses integer two-bit weights, integer eight-bit activations, and E8M0
block scales, with eight exponent bits and no mantissa bits.
Changing from G128 to this G32 configuration changes both group
size and scale representation. It is not an isolated scale-format ablation,
a floating-point FP4 result, or a kernel-speed measurement. No single-pass
weight-activation comparator is included.


\clearpage
\section*{NeurIPS Paper Checklist}

\begin{enumerate}
\item \textbf{Claims}
\item[] Question: Do the main claims made in the abstract and introduction accurately reflect the paper's contributions and scope?
\item[] Answer: \answerYes{}
\item[] Justification: The method-class result is explicitly descriptive and regime-aligned. Matched-width, scale, task-targeting, and overlap limits are stated in \cref{sec:intro,sec:results,sec:limitations}. No isolated causal mechanism is claimed.

\item \textbf{Limitations}
\item[] Question: Does the paper discuss the limitations of the work performed by the authors?
\item[] Answer: \answerYes{}
\item[] Justification: \Cref{sec:limitations} discusses comparison confounds, limited model and task coverage, unvalidated recipe choices, and unresolved overlap effects. Method-specific uncertainty is reported in \cref{sec:results}.

\item \textbf{Theory assumptions and proofs}
\item[] Question: For each theoretical result, does the paper provide the full set of assumptions and a complete (and correct) proof?
\item[] Answer: \answerNA{}
\item[] Justification: The paper reports empirical results.

\item \textbf{Experimental result reproducibility}
\item[] Question: Does the paper fully disclose all the information needed to reproduce the main experimental results of the paper to the extent that it affects the main claims and/or conclusions of the paper (regardless of whether the code and data are provided or not)?
\item[] Answer: \answerYes{}
\item[] Justification: \Cref{app:reproduction} describes calibration construction, paired quantization, checkpoint evaluation, and result aggregation. \Cref{sec:setup,app:implementation,app:calibration_accounting,app:fragcal_script} specify the models, method settings, seeds, task metrics, sample accounting, and construction code needed to reproduce the main comparison protocol.

\item \textbf{Open access to data and code}
\item[] Question: Does the paper provide open access to the data and code, with sufficient instructions to faithfully reproduce the main experimental results, as described in supplemental material?
\item[] Answer: \answerNo{}
\item[] Justification: The supplementary material provides the construction script and protocol documentation, but not the complete method-specific XPU implementations and execution environments needed to reproduce the main quantization and evaluation results.

\item \textbf{Experimental setting/details}
\item[] Question: Does the paper specify all the training and test details (e.g., data splits, hyperparameters, how they were chosen, type of optimizer) necessary to understand the results?
\item[] Answer: \answerYes{}
\item[] Justification: \Cref{sec:setup,app:fragcal,app:implementation,app:calibration_accounting} describe models, tasks, metrics, calibration splits and selection, seeds, widths, group sizes, optimization settings, and loader-count differences needed to interpret the comparisons. These details do not constitute a complete executable reproduction package.

\newpage
\item \textbf{Experiment statistical significance}
\item[] Question: Does the paper report error bars suitably and correctly defined or other appropriate information about the statistical significance of the experiments?
\item[] Answer: \answerYes{}
\item[] Justification: The main comparisons report model-level 95\% $t$ intervals and exact sign-flip tests, with seeds averaged within models (\cref{sec:setup,sec:results,sec:contamination}). Paired-seed and hierarchical-bootstrap intervals are identified separately; related-model and small-panel limitations are explicit. Trajectory summaries report across-seed standard deviations with divisor $n=4$ (\cref{app:c3_trajectory}).

\item \textbf{Experiments compute resources}
\item[] Question: For each experiment, does the paper provide sufficient information on the computer resources (type of compute workers, memory, time of execution) needed to reproduce the experiments?
\item[] Answer: \answerYes{}
\item[] Justification: \Cref{app:hardware} specifies the accelerator hardware and memory, representative per-run quantization times, evaluation-time ranges, and an author-estimated overall research budget of approximately 3,000 PVC tile-hours.

\item \textbf{Code of ethics}
\item[] Question: Does the research conducted in the paper conform, in every respect, with the NeurIPS Code of Ethics \url{https://neurips.cc/public/EthicsGuidelines}?
\item[] Answer: \answerYes{}
\item[] Justification: The authors confirm compliance with the NeurIPS Code of Ethics. The study involves no human participants; \cref{sec:conclusion} discusses societal and dual-use considerations.

\item \textbf{Broader impacts}
\item[] Question: Does the paper discuss both potential positive societal impacts and negative societal impacts of the work performed?
\item[] Answer: \answerYes{}
\item[] Justification: \Cref{sec:conclusion} discusses reduced deployment costs and resource use alongside dual-use risks and selective capability preservation or degradation.

\item \textbf{Safeguards}
\item[] Question: Does the paper describe safeguards that have been put in place for responsible release of data or models that have a high risk for misuse (e.g., pre-trained language models, image generators, or scraped datasets)?
\item[] Answer: \answerNA{}
\item[] Justification: The supplementary assets consist of code and protocol documentation; no datasets or model weights are included.

\item \textbf{Licenses for existing assets}
\item[] Question: Are the creators or original owners of assets (e.g., code, data, models), used in the paper, properly credited and are the license and terms of use explicitly mentioned and properly respected?
\item[] Answer: \answerYes{}
\item[] Justification: Sources are credited, and \cref{app:asset_terms} and the supplementary terms inventory identify their licenses and model-specific conditions. The authors confirm compliance with the applicable licenses and terms; no license conflicts were identified. The supplementary release contains code and documentation, with no datasets or model weights.

\newpage
\item \textbf{New assets}
\item[] Question: Are new assets introduced in the paper well documented and is the documentation provided alongside the assets?
\item[] Answer: \answerYes{}
\item[] Justification: The construction code and protocol are provided as supplementary files under Apache-2.0, with the full license text, usage instructions, dependencies, limitations, and source-terms inventory. \Cref{app:fragcal_script} provides source revisions and artifact checksums.

\item \textbf{Crowdsourcing and research with human subjects}
\item[] Question: For crowdsourcing experiments and research with human subjects, does the paper include the full text of instructions given to participants and screenshots, if applicable, as well as details about compensation (if any)?
\item[] Answer: \answerNA{}
\item[] Justification: The reported study does not recruit human participants.

\item \textbf{Institutional review board approvals or equivalent for research with human subjects}
\item[] Question: Does the paper describe potential risks incurred by study participants, whether such risks were disclosed to the subjects, and whether Institutional Review Board (IRB) approvals (or an equivalent approval/review based on the requirements of your country or institution) were obtained?
\item[] Answer: \answerNA{}
\item[] Justification: No human-subject experiment is reported.

\item \textbf{Declaration of LLM usage}
\item[] Question: Does the paper describe the usage of LLMs if it is an important, original, or non-standard component of the core methods in this research? Note that if the LLM is used only for writing, editing, or formatting purposes and does not impact the core methodology, scientific rigor, or originality of the research, declaration is not required.
\item[] Answer: \answerNA{}
\item[] Justification: LLMs are the models evaluated in this study; no LLM-based procedure is introduced as a component of the quantization or calibration method.
\end{enumerate}

\end{document}